\pdfoutput=1
\documentclass[letterpaper]{article} 
\usepackage{aaai23}
\usepackage{times}  
\usepackage{helvet}  
\usepackage{courier}  
\usepackage[hyphens]{url}  
\usepackage{graphicx} 
\usepackage{natbib}  
\usepackage{caption} 
\usepackage{times}  
\usepackage{helvet}  
\usepackage{courier}  
\usepackage[hyphens]{url}  
\usepackage{graphicx} 
\usepackage{natbib}  
\usepackage{caption} 
\DeclareCaptionStyle{ruled}{labelfont=normalfont,labelsep=colon,strut=off} 
\usepackage{algorithm}
\usepackage{algorithmic}
\usepackage{amsmath}
\usepackage{threeparttable}
\usepackage{booktabs}
\usepackage{multirow}
\usepackage{amsfonts}
\usepackage{subcaption}
\usepackage{color}
\usepackage{makecell}

\usepackage{algorithm}
\usepackage{algorithmic}

\usepackage{xspace}
\newcommand{\method}{CoDA\xspace} 
\usepackage{newfloat}
\usepackage{listings}
\DeclareCaptionStyle{ruled}{labelfont=normalfont,labelsep=colon,strut=off} 
\floatstyle{ruled}
\newfloat{listing}{tb}{lst}{}
\floatname{listing}{Listing}
\title{Correspondence-Free Domain Alignment for Unsupervised Cross-Domain Image Retrieval} 
\author{
    Xu Wang\textsuperscript{\rm 1}, Dezhong Peng\textsuperscript{\rm 1,\rm 3,\rm 4}, Ming Yan\textsuperscript{\rm 2}, Peng Hu\textsuperscript{\rm 1}\thanks{Corresponding author.} 
}
\affiliations{
    \textsuperscript{\rm 1}College of Computer Science, Sichuan University, Chengdu, China\\
    \textsuperscript{\rm 2}Centre for Frontier AI Research (CFAR), A*STAR, Singapore\\

    wangxu.scu@gmail.com, pengdz@scu.edu.cn, yanmingtop@gmail.com, penghu.ml@gmail.com
    
}

\begin{document}

\maketitle

\begin{abstract}
Cross-domain image retrieval aims at retrieving images across different domains to excavate cross-domain classificatory or correspondence relationships. This paper studies a less-touched problem of cross-domain image retrieval, \textit{i.e.}, unsupervised cross-domain image retrieval, considering the following practical assumptions: (i) no correspondence relationship, and (ii) no category annotations. It is challenging to align and bridge distinct domains without cross-domain correspondence. To tackle the challenge, we present a novel Correspondence-free Domain Alignment (\method) method to effectively eliminate the cross-domain gap through In-domain Self-matching Supervision (ISS) and Cross-domain Classifier Alignment (CCA). To be specific, ISS is presented to encapsulate discriminative information into the latent common space by elaborating a novel self-matching supervision mechanism. To alleviate the cross-domain discrepancy, CCA is proposed to align distinct domain-specific classifiers. Thanks to the ISS and CCA, our method could encode the discrimination into the domain-invariant embedding space for unsupervised cross-domain image retrieval. To verify the effectiveness of the proposed method, extensive experiments are conducted on four benchmark datasets compared with six state-of-the-art methods.
\end{abstract}

\section{Introduction}
With the rapid growth of images collected from many diverse sources (\textit{e.g.}, viewpoints, lightning, artistic styles, and photograph) on the Internet, there are growing demands to develop various applications on different domains, such as domain adaptation~\cite{li2021bi,singh2021clda,zhu2019aligning}, cross-domain clustering~\cite{li2021cross}, and cross-domain image retrieval~(CIR)~\cite{huang2015cross,wang2019advcae,paul2021universal,wang2022dsc3l,hu2022feature}. In these applications, CIR has attracted more and more attention in recent years for its flexible retrieval ways and achieved great success in numerous application scenarios, \textit{e.g.}, surveillance, mobile product image search~\cite{shen2012mobile}. Given a query image, CIR aims to correctly retrieve relevant images across distinct domains, which are with similar visual information or the same semantics.
However, it is challenging to retrieve images across diverse domains due to the inconsistent image distributions, namely the so-called ``domain gap'' or ``cross-domain gap''~\cite{nam2021reducing}.

\begin{figure}[tb]
\centering
    \includegraphics[width=0.48\textwidth]{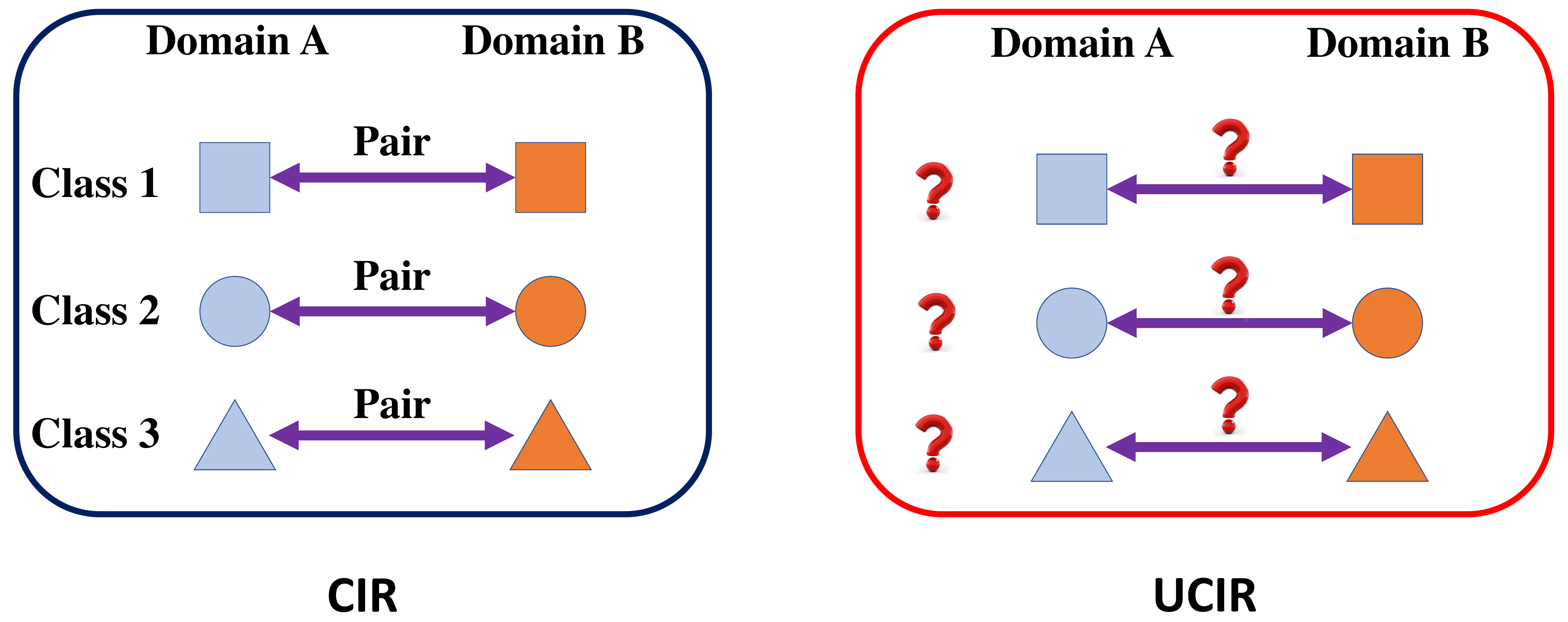}
    \caption{Comparison of cross-domain image retrieval (CIR) and unsupervised cross-domain image retrieval (UCIR). Compared with CIR, there is no category annotation or correspondence relationship in UCIR.}\label{fig:UCIR}
\end{figure}

To bridge the domain gap, extensive efforts have been devoted to learning common representations from different domains~\cite{sangkloy2016sketchy,yu2016sketch,sain2021stylemeup,fuentes2021sketch}. 
Although the existing cross-domain image retrieval methods have achieved promising performance, they implicitly assume that the multi-domain training data are annotated and aligned well. In practice, however, it is extremely expensive and even impossible to label multiple large-scale domains. 
To alleviate the high labeling cost, one advisable solution is to design an unsupervised cross-domain learning paradigm to learn from a large number of low-cost and highly accessible unlabeled data. Obviously, compared with CIR, unsupervised cross-domain image retrieval (UCIR) is more challenging due to unavailable category and correspondence information as shown in Figure \ref{fig:UCIR}. Such a UCIR problem is barely touched so far, to the best of our knowledge. 

To overcome the challenges, this paper proposes a novel approach dubbed \textbf{Co}rrespondence-free \textbf{D}omain \textbf{A}lignment~(\textbf{\method}), which unifies the In-domain Self-matching Supervision (ISS) and Cross-domain Classifier Alignment (CCA) to achieve unsupervised cross-domain image retrieval. Specifically, ISS employs a novel self-matching supervision mechanism to encapsulate the discriminative information into the shared embedding space. Meanwhile, different from previous works which mainly focus on the distribution alignment,
CCA enforces the predictions by different domain-specific classifiers to be consistent to minimize the cross-domain discrepancy, thus learning the domain-aligned and domain-invariant representations. 

The main novelties and contributions of this work are summarized as follows:

\begin{itemize}
    \item We propose a novel method called \textbf{Co}rrespondence-free \textbf{D}omain \textbf{A}lignment~(\textbf{\method}) to tackle a less-touched problem, \textit{i.e.}, unsupervised cross-domain image retrieval.
    \item A novel In-domain Self-matching Supervision module (ISS) is proposed to project the discrimination into common representations by simultaneously conducting domain-specific clustering and discriminative learning.
    \item We present a Cross-domain Classifier Alignment mechanism (CCA) to learn domain-invariant representations by minimizing the discrepancy across domain-specific classifiers.
    \item Extensive experiments are conducted on four benchmarks, demonstrating the effectiveness of the proposed approach for unsupervised cross-domain image retrieval.
\end{itemize}

\begin{figure*}[htb]
	\centering
	\includegraphics[width=0.85\textwidth]{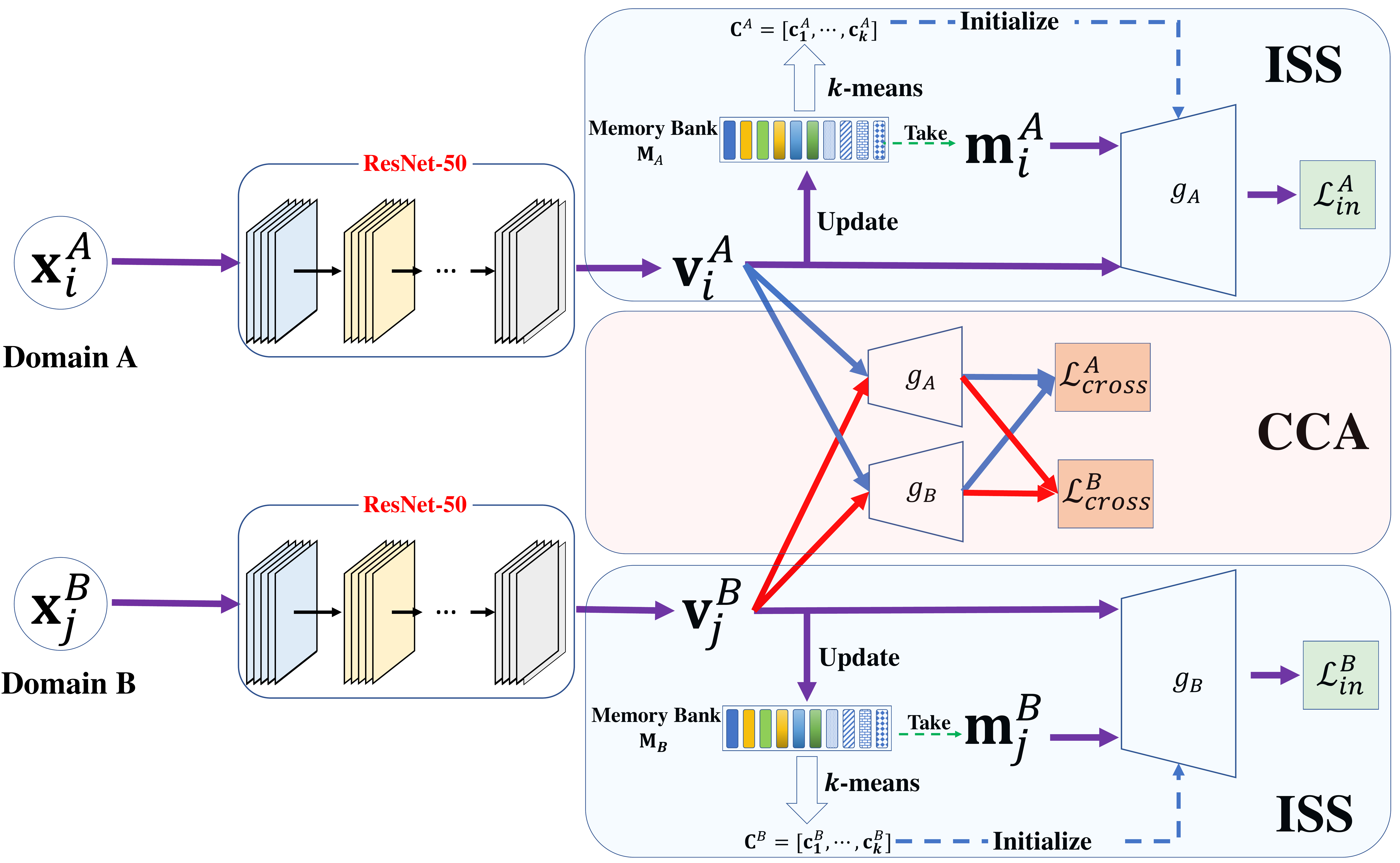}
	\caption{The pipeline of our \method for unsupervised cross-domain image retrieval. The ResNet-50 network is adopted to learn the feature embeddings for different domains. ISS performs domain-specific clustering and in-domain self-matching with the soft-label, thus encapsulating the discrimination into the embedding space. CCA minimizes cross-domain discrepancy by enforcing the predictions of different domain-specific classifiers to be consistent.}\label{fig:arch}
\end{figure*}

\section{Related Work}
\subsection{Unsupervised Domain Adaptation}
Unsupervised Domain Adaptation~(UDA) aims to transfer knowledge learned from a source domain with fully-annotated training examples to a target domain with unlabeled data only. The key challenge of UDA is having to counter the inconsistent distributions across different domains, namely the cross-domain gap. To eliminate the domain gap, discrepancy-based methods are proposed to minimize the Maximum Mean Discrepancy~(MMD)~\cite{wynne2022mmd} or Joint MMD~\cite{long2017deep} across domains. Motivated by the Generative Adversarial Networks~(GANs)~\cite{goodfellow2014gan}, several GAN-based methods~\cite{tzeng2017adversarial,hoffman2018cycada} are also presented to perform domain alignment in feature space. Besides, \citet{saito2019semi} propose a novel minimax entropy approach that adversarially achieves domain adaptation. Nevertheless, these UDA methods have supervision on the source domain. Nevertheless, these UDA methods implicitly assume that the source domain is annotated well, which inevitably increases the labeling cost. To tackle this problem, a novel problem is studied in the paper, \textit{i.e.}, the unsupervised cross-domain image retrieval, where both source and target domains are unlabeled.

\subsection{Cross-Domain Image Retrieval}
As an extensively studied task in computer vision, content-based image retrieval~\cite{datta2008image} has been widely explored, where the query and database are constrained to the same domain. In practice, however, we often require retrieving related images across diverse domains, \textit{i.e.}, cross-domain image retrieval (CIR). For example, in online shopping, we need to search for products using images captured by smartphones as the query~\cite{huang2015cross}, wherein query and database exist in distinct domains. 
Compared with content-based image retrieval, CIR is more challenging due to the domain gap. To bridge the domain gap, several works exploit the category information~\cite{sangkloy2016sketchy} for discriminative feature extraction or cross-domain pairing by minimizing triplet~\cite{yu2016sketch} and HOLEF~\cite{song2017deep} loss. However, the cross-domain correspondence utilized in these works is labor-intensive, which severely restricts their applications. In this paper, we focus on a more challenging setting, \textit{i.e.}, unsupervised cross-domain image retrieval, which is first introduced in a recent work ~\cite{kim2021cds} but is still barely touched so far. Different from this work which trains models by contrasting instance-instance pairs, ours employs multi-domain classifiers to capture discrimination and enforce their prediction consistency to alleviate the cross-domain discrepancy, thus embracing better performance.

\section{Methodology}
\subsection{Problem Statement}
We first give the formal definition of the unsupervised cross-domain retrieval task. Given two sets of unlabeled training images $\mathcal{D}_A = \{(\mathbf{x}_i^A)\}_{i=1}^{N_A}$ from domain $A$, and $\mathcal{D}_B = \{(\mathbf{x}_j^B)\}_{j=1}^{N_B}$ from another different domain $B$, our goal is to learn an effective feature extractor to transform images from different domains into a common embedding space, where the features are discriminative for cross-domain similarity measuring. During the training stage, the training set is utilized to bridge the domain gap. While during testing, a query image $\mathbf{x}_i^A \in \mathcal{D}_A$ with its category label $y_i$ is given, the ultimate goal is to correctly retrieve all semantically similar images in $\mathcal{D}_B$, \textit{i.e.}, all $\mathbf{x}_k^B \in \mathcal{D}_B$ with category label $y_k=y_i$. Under the unsupervised setting, $\mathcal{D}_A$ and $\mathcal{D}_B$ share the same categories, but there are unavailable category annotations or correspondence relationships between $\mathcal{D}_A$ and $\mathcal{D}_B$. 
This increases the challenge of how to learn a common space from unlabeled and unaligned data for cross-domain retrieval. 
\subsection{Overview}
To tackle the above-mentioned challenge, we propose a novel approach named Correspondence-free Domain Alignment~(\method). We firstly adopt a Convolutional Neural Network~(CNN) backbone $f(\cdot; \boldsymbol{\Theta})$ pretrained on ImageNet to extract the features. Formally, given an image point $\mathbf{x}_i$, the  embedded feature $\mathbf{v}_i$ could be obtained by: 
\begin{equation}
\label{u}
    \mathbf{v}_i = f(\mathbf{x}_i; \boldsymbol{\Theta}) \in \mathbb{R}^{L},
\end{equation}
where $\boldsymbol{\Theta}$ is the parameter set of CNN. $L$ denotes the dimensionality of the common space. 

As illustrated in Figure \ref{fig:arch}, our \method contains two parts:

\textbf{(1) In-domain Self-matching Supervision (ISS).} By performing domain-specific clustering and in-domain self-matching with the soft label, ISS could encapsulate the discriminative information of image data into the embedding space.

\textbf{(2) Cross-domain Classifier Alignment (CCA).} By minimizing the discrepancy among different domain-specific classifiers, CCA
achieves domain alignment through enforcing the predictions by different domain-specific classifiers to be consistent. 

To learn discriminative common representations, our \method unifies ISS and CCA to encapsulate discrimination while eliminating the cross-domain discrepancy in the latent common space. The overall objective function could be formulated as:
\begin{equation}\label{eq:L_join}
    \mathcal{L}_{join} = \mathcal{L}_{in} + \lambda \mathcal{L}_{cross},
\end{equation}
where $\mathcal{L}_{in}$ and $\mathcal{L}_{cross}$ are the objectives of ISS and CCA, respectively. $\lambda$ is the parameter to balance the contribution of $\mathcal{L}_{cross}$. 
Due to the randomness of clustering, we perform $k$-means on the samples $R$ times with different numbers of clusters $\{k_r\}_{r=1}^{R}$ for relatively stable results. In summary, the objective function of \method can be rewritten as:
\begin{equation}\label{loss}
    \mathcal{L} = \frac{1}{R}\sum_{r=1}^{R}\mathcal{L}^{(r)}_{join}
\end{equation}
To train the proposed method, we adopt a gradient descent optimizer to minimize the objective function in a batch-by-batch manner. Algorithm~\ref{alg:optim_proc} briefly summarizes the optimization procedure of the proposed \method approach.

\subsection{In-Domain Self-Matching Supervision}
In UCIR task, it is important to learn discriminative features. 
Intuitively, we expect that the features within the same cluster stay close while those in different clusters are apart from each other. 
To this end, we propose a novel In-domain Self-matching Supervision (ISS) module. There are two key steps in our ISS: (1) domain-specific clustering and (2) in-domain self-matching with the soft label. 

%

\subsubsection{(1) Domain-specific Clustering.} 
We firstly maintain two memory banks $\mathbf{M}_A$ and $\mathbf{M}_B$ for $\mathcal{D}_A$ and $\mathcal{D}_B$ respectively:
\begin{equation}
    \mathbf{M}_A=[\mathbf{m}_1^A,\cdots,\mathbf{m}_{N_A}^A], \mathbf{M}_B=[\mathbf{m}_1^B,\cdots,\mathbf{m}_{N_B}^B].
\end{equation}
The memory banks are initialized with the features extracted by $f(\cdot; \boldsymbol{\Theta})$. During training, the features in memory banks $\mathbf{M}_A$ and $\mathbf{M}_B$ are updated with a momentum $\eta$ after every batch:
\begin{equation}\label{memorybank_update}
\begin{split}
    \mathbf{m}_{i}^A &= \eta\mathbf{m}_{i}^A + (1-\eta)\mathbf{v}_{i}^A,\\
    \mathbf{m}_{j}^B &= \eta\mathbf{m}_{j}^B + (1-\eta)\mathbf{v}_{j}^B.
\end{split}
\end{equation}
To reduce the high memory cost of gradient computation, the memory banks are updated without producing gradients.
After initialization, we then respectively perform $k$-means clustering on $\mathbf{M}_A$ and $\mathbf{M}_B$ to obtain domain-specific centroids $\mathbf{C}^A=[\mathbf{c}_1^A,\cdots,\mathbf{c}_k^A]$ and $\mathbf{C}^B=[\mathbf{c}_1^B,\cdots,\mathbf{c}_k^B]$. Note that, we use the union of the features from two domains and perform $k$-means to get global clustering centroids, which are leveraged as the initialized centroids of domain-specific clustering. Meanwhile, different from Deep Clustering~\cite{caron2018deepclustering} which conducts $k$-means every epoch, our method only conducts $k$-means clustering once in the first epoch, thus embracing higher efficiency.

\subsubsection{(2) In-domain Self-matching with Soft Label.}
In order to learn discriminative representations for each domain, we design two classifiers $\sigma(g_A(\cdot))$ and $\sigma(g_B(\cdot))$ for $\mathcal{D}_A$ and $\mathcal{D}_B$ respectively, where $\sigma(\cdot)$ denotes the softmax function, $g_A(\cdot)$ and $g_B(\cdot)$ are two linear layers with weights $\mathbf{W}_A$ and $\mathbf{W}_B$, which are respectively initialized with domain-specific centroids $\mathbf{C}^A$ and $\mathbf{C}^B$. Since there are unavailable labels in unsupervised cross-domain image retrieval, we need a self-labeling mechanism to assign the labels automatically. 
During the self-labeling, the memory bank $\mathbf{M}_A$ is initialized with $\mathbf{v}_i^A$.
Based on the consistency regularization~\cite{bachman2014consistency_regularization,sohn2020fixmatch}, which holds that the model
should output similar predictions when fed augmented versions of the same image, we regard the memory feature $\mathbf{m}_i^A$ as an augmentation of $\mathbf{v}_i^A$, proposing to achieve in-domain self-matching via the loss function:
\begin{equation}
    \mathcal{L}_{in}^A = \frac{1}{N_A}\sum_{i=1}^{N_A}\text{H}\left( \sigma \left( \frac{g_A(\mathbf{m}_i^A)}{\tau} \right), \sigma \left( g_A(\mathbf{v}_i^A \right) \right),
\end{equation}
where $\sigma(\frac{g_A(\mathbf{m}_i^A)}{\tau})$ is the soft label of image $\mathbf{x}_i^A$, and $\text{H}(p, q)$ denotes the cross-entropy between two probability distributions $p$ and $q$, and $\tau$ is the temperature parameter.
Similarly, we have the loss for domain $\mathcal{D}_B$ as $\mathcal{L}_{in}^B$:
\begin{equation}
    \mathcal{L}_{in}^B = \frac{1}{N_B}\sum_{j=1}^{N_B}\text{H}\left( \sigma \left( \frac{g_B(\mathbf{m}_j^B)}{\tau} \right), \sigma \left( g_B(\mathbf{v}_j^B \right) \right).
\end{equation}
Then the loss of ISS can be written as:
\begin{equation}\label{eq:L_in}
    \mathcal{L}_{in} = \mathcal{L}_{in}^A + \mathcal{L}_{in}^B.
\end{equation}

\subsection{Cross-Domain Classifier Alignment}
With ISS, the model is supposed to learn discriminative features for each domain. However, it ignores the domain invariance, which is another important demand for cross-domain image retrieval. In an unsupervised setting, it is challenging to learn domain-invariant features since there is no correspondence between different domains. To encourage domain-aligned as well as domain-invariant features across different domains, we propose a Cross-domain Classifier Alignment mechanism (CCA), which minimizes the discrepancy between different domain-specific classifiers.

For our proposal, we hold that the classifiers which are trained on different domains have a disagreement on the predictions of the same feature. The disagreement is especially obvious when the features are near the class boundaries. To learn domain-invariant features across different domains, the predictions by domain-specific classifiers of the same image should be consistent. Thus, the discrepancy across different domain-specific classifiers is supposed to be minimized. 

Concretely, take the feature $\mathbf{v}_i^A$ as example, which is feed into the domain-specific classifiers $g_A(\cdot)$ and $g_B(\cdot)$ to get logits. Then we employ the mean absolute values of the difference between the logits of different domain-specific classifiers as the cross-domain alignment loss:

\begin{equation}
    \mathcal{L}_{cross}^{A}= \frac{1}{N_A}\sum_{i=1}^{N_A}\left| g_A(\mathbf{v}_i^A) - g_B(\mathbf{v}_i^A) \right|.
\end{equation}
Similarly, we have the cross-domain alignment loss for $\mathbf{v}_j^B$ as follows:
\begin{equation}
    \mathcal{L}_{cross}^{B}= \frac{1}{N_B}\sum_{j=1}^{N_B}\left| g_A(\mathbf{v}_j^B) - g_B(\mathbf{v}_j^B) \right|.
\end{equation}
Finally, the loss of CCA can be written as:
\begin{equation}\label{eq:L_cross}
    \mathcal{L}_{cross} = \mathcal{L}_{cross}^A + \mathcal{L}_{cross}^B.
\end{equation}

\begin{algorithm}[t]
		\caption{Optimization procedure of \method}
		\label{alg:optim_proc}
		\renewcommand{\algorithmicrequire}{\textbf{Input:}}
		\renewcommand{\algorithmicensure}{\textbf{Output:}}
		\begin{algorithmic}[1]
			\REQUIRE The training dataset
			$\mathcal{D}_A = \{(\mathbf{x}_i^A)\}_{i=1}^{N_A}$ from domain $A$, and $\mathcal{D}_B = \{(\mathbf{x}_j^B)\}_{j=1}^{N_B}$ from domain $B$, the dimensionality of the common space $L$, memory bank update momentum $\eta$, batch size $n_{b}$, balance parameter $\lambda$, maximal epoch number $N_e$, numbers of clustering $\{k_r\}_{r=1}^{R}$, temperature parameter $\tau$, and initial learning rate $\alpha$. \\
            \STATE{Calculate the features for all images from both domains by using the backbone network $f(\cdot, \boldsymbol{\Theta})$ according to Equation (\ref{u}).}
			\STATE{Initialize memory banks $\mathbf{M}_A$ and $\mathbf{M}_B$ with the calculated features.}
			\STATE{Perform domain-specific clustering and initialize the domain-specific classifiers with domain-specific centroids.}
			\FOR{$1,2,\cdots, N_e$}
			\REPEAT
			\STATE{Randomly select $n_{b}$ images from $\mathcal{D}_A$ and $n_{b}$ images from $\mathcal{D}_B$ to construct mini-batch data.}
			\STATE{Calculate the representations for all images of the mini-batch by using the backbone network $f(\cdot, \boldsymbol{\Theta})$ according to Equation (\ref{u}).}
			\STATE{Compute $\mathcal{L}_{in}$ and $\mathcal{L}_{cross}$ according to Equations (\ref{eq:L_in}) and (\ref{eq:L_cross}) on the mini-batch, respectively.}
			\STATE{Update network parameters $\boldsymbol{\Psi}=\{\boldsymbol{\Theta},\mathbf{W}_A,\mathbf{W}_B\}$ by minimizing $\mathcal{L}$ in Equation~(\ref{loss}) with descending their stochastic gradient.}
			\STATE{Update memory banks according to Equation~(\ref{memorybank_update}).}
            \UNTIL{all images are selected}
            \ENDFOR
			\ENSURE Optimized network parameters $\{\boldsymbol{\Theta},\mathbf{W}_A,\mathbf{W}_B\}$.
		\end{algorithmic}
\end{algorithm}

\begin{table*}[h]
\centering
\resizebox{\linewidth}{!}{
\begin{tabular}{cccccccccccccccc}
\toprule
\hline
\multicolumn{2}{c}{\multirow{3}{*}{Method}}           & \multicolumn{14}{c}{Cross-domain Retrieval Task on Office31 and Adaptiope datasets}  \\ \cline{3-16} 
\multicolumn{2}{c}{}                                  & \multicolumn{7}{c|}{Office31}                                 & \multicolumn{7}{c}{Adaptiope}   \\ \cline{3-16} 
\multicolumn{2}{c}{}                                  & \multicolumn{1}{c}{A-D}  & \multicolumn{1}{c}{A-W}  & \multicolumn{1}{c}{D-A} & \multicolumn{1}{c}{D-W} & \multicolumn{1}{c}{W-A} & \multicolumn{1}{c}{W-D} & \multicolumn{1}{c|}{Avg} & \multicolumn{1}{c}{P-R} & \multicolumn{1}{c}{P-S} & \multicolumn{1}{c}{R-P} & \multicolumn{1}{c}{R-S} & \multicolumn{1}{c}{S-P} & \multicolumn{1}{c}{S-R} & Avg \\ \hline
\multicolumn{1}{l}{ImageNet}     &   /   & \multicolumn{1}{c}{0.569}      & \multicolumn{1}{c}{0.500}     & \multicolumn{1}{c}{0.617}    & \multicolumn{1}{c}{0.816}    &  \multicolumn{1}{c}{0.552}   & \multicolumn{1}{c}{0.801} & \multicolumn{1}{c|}{0.643}    & \multicolumn{1}{c}{0.400}    & \multicolumn{1}{c}{0.191}    & \multicolumn{1}{c}{0.395}    & \multicolumn{1}{c}{0.137}    & \multicolumn{1}{c}{0.222}    & \multicolumn{1}{c}{0.165}    &   \multicolumn{1}{c}{0.252}  \\ 

\multicolumn{1}{l}{\multirow{2}{*}{MMD~\cite{wynne2022mmd}}}       
& Best & \multicolumn{1}{c}{0.426}      & \multicolumn{1}{c}{0.356}     & \multicolumn{1}{c}{0.524}    & \multicolumn{1}{c}{0.716}    &     
\multicolumn{1}{c}{0.450}    & \multicolumn{1}{c}{0.709} & \multicolumn{1}{c|}{0.529} & \multicolumn{1}{c}{0.274} & \multicolumn{1}{c}{0.100} & \multicolumn{1}{c}{0.244} & \multicolumn{1}{c}{0.066} & \multicolumn{1}{c}{0.093} & \multicolumn{1}{c}{0.090} & \multicolumn{1}{c}{0.145}\\                                
& Last & \multicolumn{1}{c}{0.146}      & \multicolumn{1}{c}{0.126}     & \multicolumn{1}{c}{0.460}    & \multicolumn{1}{c}{0.673}    &   \multicolumn{1}{c}{0.319}  & \multicolumn{1}{c}{0.636}    & \multicolumn{1}{c|}{0.393} & \multicolumn{1}{c}{0.016} & \multicolumn{1}{c}{0.014} & \multicolumn{1}{c}{0.013} & \multicolumn{1}{c}{0.013} & \multicolumn{1}{c}{0.017} & \multicolumn{1}{c}{0.014} & \multicolumn{1}{c}{0.015}\\

\multicolumn{1}{l}{\multirow{2}{*}{SimCLR~\cite{chen2020simple}}}       
& Best & \multicolumn{1}{c}{0.544}      & \multicolumn{1}{c}{0.496}     & \multicolumn{1}{c}{0.617}    & \multicolumn{1}{c}{0.819}    &     
\multicolumn{1}{c}{0.543}    & \multicolumn{1}{c}{0.812} & \multicolumn{1}{c|}{0.638} & \multicolumn{1}{c}{0.362} & \multicolumn{1}{c}{0.202} & \multicolumn{1}{c}{0.343} & \multicolumn{1}{c}{0.135} & \multicolumn{1}{c}{0.211} & \multicolumn{1}{c}{0.156} & \multicolumn{1}{c}{0.235}\\                                
& Last & \multicolumn{1}{c}{0.540}      & \multicolumn{1}{c}{0.473}     & \multicolumn{1}{c}{0.614}    & \multicolumn{1}{c}{0.807}    &   \multicolumn{1}{c}{0.535}  & \multicolumn{1}{c}{0.804}    & \multicolumn{1}{c|}{0.629} & \multicolumn{1}{c}{0.254} & \multicolumn{1}{c}{0.135} & \multicolumn{1}{c}{0.248} & \multicolumn{1}{c}{0.091} & \multicolumn{1}{c}{0.160} & \multicolumn{1}{c}{0.117} & \multicolumn{1}{c}{0.168}\\

\multicolumn{1}{l}{\multirow{2}{*}{InstDis~\cite{wu2018unsupervised}}}              
& Best & \multicolumn{1}{c}{0.509}      & \multicolumn{1}{c}{0.452}     & \multicolumn{1}{c}{0.640}    & \multicolumn{1}{c}{0.847}    &     
\multicolumn{1}{c}{0.570}    & \multicolumn{1}{c}{0.812} & \multicolumn{1}{c|}{0.638} & \multicolumn{1}{c}{0.429} & \multicolumn{1}{c}{0.222} & \multicolumn{1}{c}{0.418} & \multicolumn{1}{c}{0.164} & \multicolumn{1}{c}{0.241} & \multicolumn{1}{c}{0.183} & \multicolumn{1}{c}{0.276} \\                                
& Last & \multicolumn{1}{c}{0.329}      & \multicolumn{1}{c}{0.263}     & \multicolumn{1}{c}{0.520}    & \multicolumn{1}{c}{0.704}    &   \multicolumn{1}{c}{0.434}  & \multicolumn{1}{c}{0.602}    & \multicolumn{1}{c|}{0.475} & \multicolumn{1}{c}{0.331} & \multicolumn{1}{c}{0.151} & \multicolumn{1}{c}{0.325} & \multicolumn{1}{c}{0.114} & \multicolumn{1}{c}{0.170} & \multicolumn{1}{c}{0.135} & \multicolumn{1}{c}{0.204}\\ 

\multicolumn{1}{l}{\multirow{2}{*}{CDS\cite{kim2021cds}}}              
& Best & \multicolumn{1}{c}{0.667}      & \multicolumn{1}{c}{0.625}     & \multicolumn{1}{c}{0.709}    & \multicolumn{1}{c}{\underline{0.900}}    &     
\multicolumn{1}{c}{0.644}    & \multicolumn{1}{c}{0.884} & \multicolumn{1}{c|}{0.738} & \multicolumn{1}{c}{\underline{0.575}} & \multicolumn{1}{c}{\underline{0.352}} & \multicolumn{1}{c}{0.574} & \multicolumn{1}{c}{0.250} & \multicolumn{1}{c}{\underline{0.361}} & \multicolumn{1}{c}{0.254} & \multicolumn{1}{c}{\underline{0.394}}\\                                
& Last & \multicolumn{1}{c}{0.520}      & \multicolumn{1}{c}{0.478}     & \multicolumn{1}{c}{0.693}    & \multicolumn{1}{c}{0.864}    &   \multicolumn{1}{c}{0.624}  & \multicolumn{1}{c}{0.796}    & \multicolumn{1}{c|}{0.663} & \multicolumn{1}{c}{0.540} & \multicolumn{1}{c}{0.328} & \multicolumn{1}{c}{0.549} & \multicolumn{1}{c}{0.212} &\multicolumn{1}{c}{\underline{0.333}} & \multicolumn{1}{c}{0.228} & \multicolumn{1}{c}{0.365}\\ 

\multicolumn{1}{l}{\multirow{2}{*}{PCS~\cite{yue2021prototypical}}}              
& Best & \multicolumn{1}{c}{\textbf{0.727}}      & \multicolumn{1}{c}{\underline{0.707}}     & \multicolumn{1}{c}{\textbf{0.753}}    & \multicolumn{1}{c}{0.885}    &     
\multicolumn{1}{c}{\underline{0.712}}    & \multicolumn{1}{c}{\underline{0.892}} & \multicolumn{1}{c|}{\underline{0.779}} & \multicolumn{1}{c}{0.569} & \multicolumn{1}{c}{0.348} & \multicolumn{1}{c}{\textbf{0.583}} & \multicolumn{1}{c}{\underline{0.270}} & \multicolumn{1}{c}{0.337} & \multicolumn{1}{c}{\underline{0.259}} & \multicolumn{1}{c}{\underline{0.394}}\\  

& Last & \multicolumn{1}{c}{\textbf{0.711}}      & \multicolumn{1}{c}{\underline{0.692}}     & \multicolumn{1}{c}{\underline{0.742}}    & \multicolumn{1}{c}{\underline{0.875}}    &   \multicolumn{1}{c}{\underline{0.706}}  & \multicolumn{1}{c}{\underline{0.886}}    & \multicolumn{1}{c|}{\underline{0.769}} & \multicolumn{1}{c}{\underline{0.556}} & \multicolumn{1}{c}{\underline{0.340}} & \multicolumn{1}{c}{\underline{0.561}} & \multicolumn{1}{c}{\textbf{0.256}} & \multicolumn{1}{c}{0.332} & \multicolumn{1}{c}{\underline{0.244}} & \multicolumn{1}{c}{\underline{0.382}}\\ 
\hline 

\multicolumn{1}{l}{\multirow{2}{*}{\method~(ours)}}& Best & \multicolumn{1}{c}{\underline{0.717}}      & \multicolumn{1}{c}{\textbf{0.714}}    & \multicolumn{1}{c}{\underline{0.749}}    & \multicolumn{1}{c}{\textbf{0.914}}    &   \multicolumn{1}{c}{\textbf{0.731}}    &  \multicolumn{1}{c}{\textbf{0.902}} & \multicolumn{1}{c|}{\textbf{0.788}}  & \multicolumn{1}{c}{\textbf{0.598}} & \multicolumn{1}{c}{\textbf{0.376}} & \multicolumn{1}{c}{\underline{0.582}} & \multicolumn{1}{c}{\textbf{0.286}} & \multicolumn{1}{c}{\textbf{0.393}} & \multicolumn{1}{c}{\textbf{0.301}} & \multicolumn{1}{c}{\textbf{0.423}}\\

& Last & \multicolumn{1}{c}{\underline{0.709}}      & \multicolumn{1}{c}{\textbf{0.698}}     & \multicolumn{1}{c}{\textbf{0.743}}    & \multicolumn{1}{c}{\textbf{0.914}}  & \multicolumn{1}{c}{\textbf{0.721}}   &  \multicolumn{1}{c}{\textbf{0.901}} & \multicolumn{1}{c|}{\textbf{0.781}} & \multicolumn{1}{c}{\textbf{0.587}} & \multicolumn{1}{c}{\textbf{0.343}} & \multicolumn{1}{c}{\textbf{0.574}} & \multicolumn{1}{c}{\underline{0.254}} & \multicolumn{1}{c}{\textbf{0.383}}& \multicolumn{1}{c}{\textbf{0.281}} & \multicolumn{1}{c}{\textbf{0.404}}\\ 
\hline
\bottomrule
\end{tabular}
}
\caption{The mAP@All retrieval performance comparison for the proposed \method approach and other compared methods on Office31 dataset. The best and second best performance results among all methods are in bold and in underline, \textit{resp}.}\label{exp_results_office31_adaptiope}
\end{table*}

\begin{table*}[h]
\centering
\resizebox{\linewidth}{!}{
\begin{tabular}{ccccccccccccccc}
\toprule
\hline
\multicolumn{2}{c}{\multirow{2}{*}{Method}}                   & \multicolumn{12}{c}{Cross-domain Retrieval Task on ImageCLEF dataset}   \\ \cline{3-15} 
\multicolumn{2}{c}{}                                          & \multicolumn{1}{c}{B-C} & \multicolumn{1}{c}{B-I}  & \multicolumn{1}{c}{B-P} & \multicolumn{1}{c}{C-B} & \multicolumn{1}{c}{C-I} & \multicolumn{1}{c}{C-P} & \multicolumn{1}{c}{I-B} & \multicolumn{1}{c}{I-C} & \multicolumn{1}{c}{I-P} & \multicolumn{1}{c}{P-B} & \multicolumn{1}{c}{P-C} & \multicolumn{1}{c}{P-I} & Avg\\ \hline


\multicolumn{1}{l}{ImageNet}                          &  /    & \multicolumn{1}{c}{0.553}      & \multicolumn{1}{c}{0.498}     & \multicolumn{1}{c}{0.432}    & \multicolumn{1}{c}{0.538}    & \multicolumn{1}{c}{0.754}    & \multicolumn{1}{c}{0.654}    & \multicolumn{1}{c}{0.479}    & \multicolumn{1}{c}{0.742}    & \multicolumn{1}{c}{0.635}    & \multicolumn{1}{c}{0.395}    & \multicolumn{1}{c}{0.640}    &  \multicolumn{1}{c}{0.619}  &  \multicolumn{1}{c}{0.578}  \\ 

\multicolumn{1}{l}{\multirow{2}{*}{MMD~\cite{wynne2022mmd}}}               & Best & \multicolumn{1}{c}{0.496}  & \multicolumn{1}{c}{0.466} & \multicolumn{1}{c}{0.363}    & \multicolumn{1}{c}{0.504}    & \multicolumn{1}{c}{0.714}    & \multicolumn{1}{c}{0.558}    & \multicolumn{1}{c}{0.487}    & \multicolumn{1}{c}{0.705}    & \multicolumn{1}{c}{0.554}    & \multicolumn{1}{c}{0.371}    & \multicolumn{1}{c}{0.540}    &  \multicolumn{1}{c}{0.558}   & \multicolumn{1}{c}{0.526}\\ 

& Last & \multicolumn{1}{c}{0.442}  & \multicolumn{1}{c}{0.414} & \multicolumn{1}{c}{0.312}    & \multicolumn{1}{c}{0.453}    & \multicolumn{1}{c}{0.620}    & \multicolumn{1}{c}{0.476}    & \multicolumn{1}{c}{0.430}    & \multicolumn{1}{c}{0.600}    & \multicolumn{1}{c}{0.474}    & \multicolumn{1}{c}{0.323}    & \multicolumn{1}{c}{0.460}    &  \multicolumn{1}{c}{0.487}   & \multicolumn{1}{c}{0.458}   \\ 

\multicolumn{1}{l}{\multirow{2}{*}{SimCLR~\cite{chen2020simple}}}               & Best & \multicolumn{1}{c}{0.552}  & \multicolumn{1}{c}{0.508} & \multicolumn{1}{c}{0.431}    & \multicolumn{1}{c}{0.537}    & \multicolumn{1}{c}{0.790}    & \multicolumn{1}{c}{0.667}    & \multicolumn{1}{c}{0.498}    & \multicolumn{1}{c}{0.793}    & \multicolumn{1}{c}{0.647}    & \multicolumn{1}{c}{0.410}    & \multicolumn{1}{c}{0.670}    &  \multicolumn{1}{c}{0.640}   & \multicolumn{1}{c}{0.595}\\ 

& Last & \multicolumn{1}{c}{0.543}  & \multicolumn{1}{c}{0.503} & \multicolumn{1}{c}{0.426}    & \multicolumn{1}{c}{0.534}    & \multicolumn{1}{c}{0.780}    & \multicolumn{1}{c}{0.661}    & \multicolumn{1}{c}{0.485}    & \multicolumn{1}{c}{0.792}    & \multicolumn{1}{c}{0.640}    & \multicolumn{1}{c}{0.410}    & \multicolumn{1}{c}{0.669}    &  \multicolumn{1}{c}{0.631}   & \multicolumn{1}{c}{0.590}   \\ 

\multicolumn{1}{l}{\multirow{2}{*}{InstDis~\cite{wu2018unsupervised}}}               & Best & \multicolumn{1}{c}{0.521}  & \multicolumn{1}{c}{0.492} & \multicolumn{1}{c}{0.418}    & \multicolumn{1}{c}{0.530}    & \multicolumn{1}{c}{0.748}    & \multicolumn{1}{c}{0.638}    & \multicolumn{1}{c}{0.487}    & \multicolumn{1}{c}{0.718}    & \multicolumn{1}{c}{0.641}    & \multicolumn{1}{c}{0.404}    & \multicolumn{1}{c}{0.600}    &  \multicolumn{1}{c}{0.608}   & \multicolumn{1}{c}{0.567}\\ 

& Last & \multicolumn{1}{c}{0.376}  & \multicolumn{1}{c}{0.373} & \multicolumn{1}{c}{0.314}    & \multicolumn{1}{c}{0.365}    & \multicolumn{1}{c}{0.477}    & \multicolumn{1}{c}{0.383}    & \multicolumn{1}{c}{0.376}    & \multicolumn{1}{c}{0.462}    & \multicolumn{1}{c}{0.470}    & \multicolumn{1}{c}{0.318}    & \multicolumn{1}{c}{0.373}    &  \multicolumn{1}{c}{0.455}   & \multicolumn{1}{c}{0.395}   \\ 

\multicolumn{1}{l}{\multirow{2}{*}{CDS~\cite{kim2021cds}}}               & Best & \multicolumn{1}{c}{0.643}  & 
\multicolumn{1}{c}{0.632} & 
\multicolumn{1}{c}{0.532}    &
\multicolumn{1}{c}{0.643}    & 
\multicolumn{1}{c}{\underline{0.912}}    & \multicolumn{1}{c}{\underline{0.778}}    & \multicolumn{1}{c}{\underline{0.627}}    & \multicolumn{1}{c}{\textbf{0.910}}    & \multicolumn{1}{c}{\underline{0.796}}    & \multicolumn{1}{c}{0.526}    & \multicolumn{1}{c}{0.747}    &  \multicolumn{1}{c}{0.768}   & \multicolumn{1}{c}{0.709}\\ 

& Last & \multicolumn{1}{c}{0.529}  & \multicolumn{1}{c}{0.541} & \multicolumn{1}{c}{0.440}    & \multicolumn{1}{c}{0.529}    & \multicolumn{1}{c}{0.779}    & \multicolumn{1}{c}{0.664}    & \multicolumn{1}{c}{0.509}    & \multicolumn{1}{c}{0.787}    & \multicolumn{1}{c}{0.656}    & \multicolumn{1}{c}{0.394}    & \multicolumn{1}{c}{0.652}    &  \multicolumn{1}{c}{0.609}   & \multicolumn{1}{c}{0.591}   \\ 

\multicolumn{1}{l}{\multirow{2}{*}{PCS~\cite{yue2021prototypical}}}               & Best & \multicolumn{1}{c}{\underline{0.673}}  & \multicolumn{1}{c}{\underline{0.635}} & \multicolumn{1}{c}{\underline{0.534}}    & \multicolumn{1}{c}{\underline{0.688}}    & \multicolumn{1}{c}{\textbf{0.918}}    & 
\multicolumn{1}{c}{0.777}  & 
\multicolumn{1}{c}{0.619}    & \multicolumn{1}{c}{\underline{0.903}}    & \multicolumn{1}{c}{0.776}    & \multicolumn{1}{c}{\underline{0.535}}    & \multicolumn{1}{c}{\underline{0.749}}    &  \multicolumn{1}{c}{\underline{0.770}}   & \multicolumn{1}{c}{\underline{0.714}}\\ 

& Last & \multicolumn{1}{c}{\underline{0.625}}  & \multicolumn{1}{c}{\underline{0.609}} & \multicolumn{1}{c}{\underline{0.496}}    & \multicolumn{1}{c}{\underline{0.661}}    & \multicolumn{1}{c}{\textbf{0.900}}    & \multicolumn{1}{c}{\underline{0.767}}    & \multicolumn{1}{c}{\underline{0.576}}    & \multicolumn{1}{c}{\underline{0.886}}    & \multicolumn{1}{c}{\underline{0.745}}    & \multicolumn{1}{c}{\underline{0.461}}    & \multicolumn{1}{c}{\underline{0.745}}    &  \multicolumn{1}{c}{\underline{0.770}}   & \multicolumn{1}{c}{\underline{0.687}}\\ 
\hline

\multicolumn{1}{l}{\multirow{2}{*}{\method~(ours)}}              
& Best & \multicolumn{1}{c}{\textbf{0.688}}& \multicolumn{1}{c}{\textbf{0.656}}     & \multicolumn{1}{c}{\textbf{0.556}}    & \multicolumn{1}{c}{\textbf{0.705}}    & \multicolumn{1}{c}{0.901}    & \multicolumn{1}{c}{\textbf{0.785}}    & \multicolumn{1}{c}{\textbf{0.661}}    & \multicolumn{1}{c}{\textbf{0.910}}    & \multicolumn{1}{c}{\textbf{0.784}}    & \multicolumn{1}{c}{\textbf{0.552}}    & \multicolumn{1}{c}{\textbf{0.752}}    &  \multicolumn{1}{c}{\textbf{0.773}}   & \multicolumn{1}{c}{\textbf{0.727}}   \\ 

& Last & \multicolumn{1}{c}{\textbf{0.685}}& \multicolumn{1}{c}{\textbf{0.656}}     & \multicolumn{1}{c}{\textbf{0.556}}    & \multicolumn{1}{c}{\textbf{0.700}}    & \multicolumn{1}{c}{\underline{0.880}}    & \multicolumn{1}{c}{\textbf{0.774}}    & \multicolumn{1}{c}{\textbf{0.660}}    & \multicolumn{1}{c}{\textbf{0.900}}    & \multicolumn{1}{c}{\textbf{0.778}}    & \multicolumn{1}{c}{\textbf{0.550}}    & \multicolumn{1}{c}{\textbf{0.749}}    &  \multicolumn{1}{c}{\textbf{0.771}}   & \multicolumn{1}{c}{\textbf{0.721}}    \\
\hline
\bottomrule
\end{tabular}
}
\caption{The mAP@All retrieval performance comparison for the proposed \method approach and other compared methods on ImageCLEF dataset. The best and second best performance results among all methods are in bold and in underline, \textit{resp}.}\label{exp_results_image_clef}
\end{table*}


\section{Experiments}

\subsection{Datasets}
To verify the effectiveness of the proposed method, we conduct extensive experiments on four benchmark datasets, \textit{i.e.}, Office31~\cite{saenko2010adapting}, Image-CLEF~\cite{long2017deep},  OfficeHome~\cite{venkateswara2017deep}, and Adaptiope~\cite{ringwald2021adaptiope}. For each dataset, we randomly partition the data into training and test sets, with an 80-20 ratio for each category. \textbf{Office31}: This dataset consists of three real-world object domains: Amazon (A), Webcam (W), and DSLR (D). It has 4,652 images with 31 categories. We conduct six retrieval tasks: A-D, A-W, D-A, D-W, W-A, W-D. \textbf{Image-CLEF}: The dataset is a benchmark dataset for ImageCLEF 2014 domain adaptation challenge. It is composed by selecting the 12 common classes shared by four public domains: Bing (B), Caltech256 (C), ImageNet ILSVRC 2012 (I), and Pascal VOC 2012 (P). For each domain, there are 50 images in each category. We conduct 12 retrieval tasks on this dataset. \textbf{OfficeHome}: This dataset contains four domains where each domain consists of 65 categories. The four domains are Artistic images (A), Clipart (C), Product images (P), and Real-world images (R). It contains 15,500 images, with an average of around 70 images per class and a maximum of 99 images in a category. We also conduct 12 retrieval tasks for this dataset. \textbf{Adaptiope}: Adaptiope is a dataset that offers 123 classes in three different domains. The domains are Synthetic~(S), Product~(P), and Real life~(R). There are totally 36,900 images, with a maximum of 100 images in a category. We also conduct six retrieval tasks for this dataset.

\begin{table*}[h]
\centering
\resizebox{\linewidth}{!}{
\begin{tabular}{ccccccccccccccc}
\toprule
\hline
\multicolumn{2}{c}{\multirow{2}{*}{Method}}                   & \multicolumn{12}{c}{Cross-domain Retrieval Task on OfficeHome dataset}   \\ \cline{3-15} 
\multicolumn{2}{c}{}                                          & \multicolumn{1}{c}{A-C} & \multicolumn{1}{c}{A-P}  & \multicolumn{1}{c}{A-R} & \multicolumn{1}{c}{C-A} & \multicolumn{1}{c}{C-P} & \multicolumn{1}{c}{C-R} & \multicolumn{1}{c}{P-A} & \multicolumn{1}{c}{P-C} & \multicolumn{1}{c}{P-R} & \multicolumn{1}{c}{R-A} & \multicolumn{1}{c}{R-C} & \multicolumn{1}{c}{R-P} & Avg\\ \hline

\multicolumn{1}{l}{ImageNet}                          &  /    & \multicolumn{1}{c}{0.200}      & \multicolumn{1}{c}{0.325}     & \multicolumn{1}{c}{0.387}    & \multicolumn{1}{c}{0.196}    & \multicolumn{1}{c}{0.246}    & \multicolumn{1}{c}{0.268}    & \multicolumn{1}{c}{0.329}    & \multicolumn{1}{c}{0.261}    & \multicolumn{1}{c}{0.478}    & \multicolumn{1}{c}{0.362}    & \multicolumn{1}{c}{0.247}    &  \multicolumn{1}{c}{0.438}  &  \multicolumn{1}{c}{0.311}  \\ 

\multicolumn{1}{l}{\multirow{2}{*}{MMD~\cite{wynne2022mmd}}}               & Best & \multicolumn{1}{c}{0.163}  & \multicolumn{1}{c}{0.252} & \multicolumn{1}{c}{0.325}    & \multicolumn{1}{c}{0.150}    & \multicolumn{1}{c}{0.170}    & \multicolumn{1}{c}{0.208}    & \multicolumn{1}{c}{0.243}    & \multicolumn{1}{c}{0.187}    & \multicolumn{1}{c}{0.358}    & \multicolumn{1}{c}{0.295}    & \multicolumn{1}{c}{0.208}    &  \multicolumn{1}{c}{0.354}   & \multicolumn{1}{c}{0.243}\\ 

& Last & \multicolumn{1}{c}{0.083}  & \multicolumn{1}{c}{0.112} & \multicolumn{1}{c}{0.155}    & \multicolumn{1}{c}{0.075}    & \multicolumn{1}{c}{0.039}    & \multicolumn{1}{c}{0.054}    & \multicolumn{1}{c}{0.137}    & \multicolumn{1}{c}{0.046}    & \multicolumn{1}{c}{0.087}    & \multicolumn{1}{c}{0.148}    & \multicolumn{1}{c}{0.062}    &  \multicolumn{1}{c}{0.080}   & \multicolumn{1}{c}{0.090}   \\

\multicolumn{1}{l}{\multirow{2}{*}{SimCLR~\cite{chen2020simple}}}               & Best & \multicolumn{1}{c}{0.211}  & \multicolumn{1}{c}{0.314} & \multicolumn{1}{c}{0.374}    & \multicolumn{1}{c}{0.191}    & \multicolumn{1}{c}{0.229}    & \multicolumn{1}{c}{0.265}    & \multicolumn{1}{c}{0.299}    & \multicolumn{1}{c}{0.251}    & \multicolumn{1}{c}{0.459}    & \multicolumn{1}{c}{0.347}    & \multicolumn{1}{c}{0.257}    &  \multicolumn{1}{c}{0.440}   & \multicolumn{1}{c}{0.303}\\ 

& Last & \multicolumn{1}{c}{0.200}  & \multicolumn{1}{c}{0.300} & \multicolumn{1}{c}{0.364}    & \multicolumn{1}{c}{0.165}    & \multicolumn{1}{c}{0.196}    & \multicolumn{1}{c}{0.226}    & \multicolumn{1}{c}{0.282}    & \multicolumn{1}{c}{0.212}    & \multicolumn{1}{c}{0.429}    & \multicolumn{1}{c}{0.329}    & \multicolumn{1}{c}{0.222}    &  \multicolumn{1}{c}{0.424}   & \multicolumn{1}{c}{0.279}   \\

\multicolumn{1}{l}{\multirow{2}{*}{InstDis~\cite{wu2018unsupervised}}}               & Best & \multicolumn{1}{c}{0.242}  & \multicolumn{1}{c}{0.348} & \multicolumn{1}{c}{0.395}    & \multicolumn{1}{c}{0.220}    & \multicolumn{1}{c}{0.253}    & \multicolumn{1}{c}{0.277}    & \multicolumn{1}{c}{0.334}    & \multicolumn{1}{c}{0.277}    & \multicolumn{1}{c}{0.477}    & \multicolumn{1}{c}{0.365}    & \multicolumn{1}{c}{0.271}    &  \multicolumn{1}{c}{0.447}   & \multicolumn{1}{c}{0.326}\\ 

& Last & \multicolumn{1}{c}{0.221}  & \multicolumn{1}{c}{0.278} & \multicolumn{1}{c}{0.331}    & \multicolumn{1}{c}{0.173}    & \multicolumn{1}{c}{0.191}    & \multicolumn{1}{c}{0.216}    & \multicolumn{1}{c}{0.203}    & \multicolumn{1}{c}{0.205}    & \multicolumn{1}{c}{0.326}    & \multicolumn{1}{c}{0.268}    & \multicolumn{1}{c}{0.218}    &  \multicolumn{1}{c}{0.335}   & \multicolumn{1}{c}{0.247}   \\

\multicolumn{1}{l}{\multirow{2}{*}{CDS~\cite{kim2021cds}}}               & Best & \multicolumn{1}{c}{0.330}  & \multicolumn{1}{c}{0.44.5} & \multicolumn{1}{c}{0.514}    & \multicolumn{1}{c}{\underline{0.324}}    & \multicolumn{1}{c}{0.403}    & \multicolumn{1}{c}{\underline{0.418}}    & \multicolumn{1}{c}{0.452}    & \multicolumn{1}{c}{0.415}    & \multicolumn{1}{c}{0.608}    & \multicolumn{1}{c}{0.511}    & \multicolumn{1}{c}{0.420}    &  \multicolumn{1}{c}{0.588}   & \multicolumn{1}{c}{0.452}\\ 

& Last & \multicolumn{1}{c}{0.327}  & \multicolumn{1}{c}{0.438} & \multicolumn{1}{c}{0.492}    & \multicolumn{1}{c}{0.282}    & \multicolumn{1}{c}{0.360}    & \multicolumn{1}{c}{0.381}    & \multicolumn{1}{c}{0.402}    & \multicolumn{1}{c}{0.377}    & \multicolumn{1}{c}{0.542}    & \multicolumn{1}{c}{0.441}    & \multicolumn{1}{c}{0.397}    &  \multicolumn{1}{c}{0.523}   & \multicolumn{1}{c}{0.414}   \\ 

\multicolumn{1}{l}{\multirow{2}{*}{PCS~\cite{yue2021prototypical}}}               & Best & \multicolumn{1}{c}{\underline{0.343}}  & \multicolumn{1}{c}{\underline{0.463}} & \multicolumn{1}{c}{\underline{0.516}}    & \multicolumn{1}{c}{0.323}    & \multicolumn{1}{c}{\underline{0.405}}    & \multicolumn{1}{c}{0.406}    & \multicolumn{1}{c}{\underline{0.470}}    & \multicolumn{1}{c}{\underline{0.421}}    & \multicolumn{1}{c}{\underline{0.613}}    & \multicolumn{1}{c}{\underline{0.516}}    & \multicolumn{1}{c}{\underline{0.428}}    &  \multicolumn{1}{c}{\underline{0.601}}   & \multicolumn{1}{c}{\underline{0.459}}\\ 

& Last & \multicolumn{1}{c}{\underline{0.335}}  & \multicolumn{1}{c}{\underline{0.458}} & \multicolumn{1}{c}{\underline{0.513}}    & \multicolumn{1}{c}{\underline{0.306}}    & \multicolumn{1}{c}{\underline{0.390}}    & \multicolumn{1}{c}{\underline{0.398}}    & \multicolumn{1}{c}{\underline{0.452}}    & \multicolumn{1}{c}{\underline{0.412}}    & \multicolumn{1}{c}{\underline{0.605}}    & \multicolumn{1}{c}{\underline{0.492}}    & \multicolumn{1}{c}{\underline{0.405}}    &  \multicolumn{1}{c}{\underline{0.597}}   & \multicolumn{1}{c}{\underline{0.447}}\\ 
\hline

\multicolumn{1}{l}{\multirow{2}{*}{\method~(ours)}}              
& Best & \multicolumn{1}{c}{\textbf{0.347}}& \multicolumn{1}{c}{\textbf{0.496}}     & \multicolumn{1}{c}{\textbf{0.532}}    & \multicolumn{1}{c}{\textbf{0.332}}    & \multicolumn{1}{c}{\textbf{0.429}}    & \multicolumn{1}{c}{\textbf{0.447}}    & \multicolumn{1}{c}{\textbf{0.504}}    & \multicolumn{1}{c}{\textbf{0.452}}    & \multicolumn{1}{c}{\textbf{0.652}}    & \multicolumn{1}{c}{\textbf{0.531}}    & \multicolumn{1}{c}{\textbf{0.460}}    &  \multicolumn{1}{c}{\textbf{0.652}}   & \multicolumn{1}{c}{\textbf{0.486}}   \\ 
& Last & \multicolumn{1}{c}{\textbf{0.347}}& \multicolumn{1}{c}{\textbf{0.494}}     & \multicolumn{1}{c}{\textbf{0.530}}    & \multicolumn{1}{c}{\textbf{0.329}}    & \multicolumn{1}{c}{\textbf{0.421}}    & \multicolumn{1}{c}{\textbf{0.440}}    & \multicolumn{1}{c}{\textbf{0.502}}    & \multicolumn{1}{c}{\textbf{0.446}}    & \multicolumn{1}{c}{\textbf{0.648}}    & \multicolumn{1}{c}{\textbf{0.531}}    & \multicolumn{1}{c}{\textbf{0.457}}    &  \multicolumn{1}{c}{\textbf{0.650}}   & \multicolumn{1}{c}{\textbf{0.482}}    \\
\hline
\bottomrule
\end{tabular}
}
\caption{The mAP@All retrieval performance comparison for the proposed \method approach and other compared methods on OfficeHome dataset. The best and second best performance results among all methods are in bold and in underline, \textit{resp}.}\label{exp_results_officehome}
\end{table*}

\begin{table*}[h]
\centering
\resizebox{\linewidth}{!}{
\begin{tabular}{ccccccccccccccc}
\toprule
\hline
\multicolumn{2}{c}{\multirow{2}{*}{Method}}                   & \multicolumn{12}{c}{Cross-domain Retrieval Task on OfficeHome dataset}   \\ \cline{3-15} 
\multicolumn{2}{c}{}                                          & \multicolumn{1}{c}{A-C} & \multicolumn{1}{c}{A-P}  & \multicolumn{1}{c}{A-R} & \multicolumn{1}{c}{C-A} & \multicolumn{1}{c}{C-P} & \multicolumn{1}{c}{C-R} & \multicolumn{1}{c}{P-A} & \multicolumn{1}{c}{P-C} & \multicolumn{1}{c}{P-R} & \multicolumn{1}{c}{R-A} & \multicolumn{1}{c}{R-C} & \multicolumn{1}{c}{R-P} & Avg\\ \hline

\multicolumn{1}{l}{\multirow{2}{*}{\method (with $\mathcal{L}_{in}$ only)}}               & Best & \multicolumn{1}{c}{0.323}  & \multicolumn{1}{c}{0.464} & \multicolumn{1}{c}{0.518}    & \multicolumn{1}{c}{0.313}    & \multicolumn{1}{c}{0.377}    & \multicolumn{1}{c}{0.411}    & \multicolumn{1}{c}{0.476}    & \multicolumn{1}{c}{0.403}    & \multicolumn{1}{c}{0.637}    & \multicolumn{1}{c}{0.537}    & \multicolumn{1}{c}{0.422}    &  \multicolumn{1}{c}{0.626}   & \multicolumn{1}{c}{0.459}\\ 

& Last & \multicolumn{1}{c}{0.320}  & \multicolumn{1}{c}{0.463} & \multicolumn{1}{c}{0.518}    & \multicolumn{1}{c}{0.311}    & \multicolumn{1}{c}{0.374}    & \multicolumn{1}{c}{0.395}    & \multicolumn{1}{c}{0.476}    & \multicolumn{1}{c}{0.390}    & \multicolumn{1}{c}{0.628}    & \multicolumn{1}{c}{0.537}    & \multicolumn{1}{c}{0.408}    &  \multicolumn{1}{c}{0.613}   & \multicolumn{1}{c}{0.453}   \\

\multicolumn{1}{l}{\multirow{2}{*}{\method (with $\mathcal{L}_{cross}$ only)}}               & Best & \multicolumn{1}{c}{0.114}  & \multicolumn{1}{c}{0.182} & \multicolumn{1}{c}{0.238}    & \multicolumn{1}{c}{0.100}    & \multicolumn{1}{c}{0.074}    & \multicolumn{1}{c}{0.097}    & \multicolumn{1}{c}{0.179}    & \multicolumn{1}{c}{0.092}    & \multicolumn{1}{c}{0.196}    & \multicolumn{1}{c}{0.225}    & \multicolumn{1}{c}{0.099}    &  \multicolumn{1}{c}{0.149}   & \multicolumn{1}{c}{0.145}\\ 

& Last & \multicolumn{1}{c}{0.101}  & \multicolumn{1}{c}{0.141} & \multicolumn{1}{c}{0.158}    & \multicolumn{1}{c}{0.084}    & \multicolumn{1}{c}{0.041}    & \multicolumn{1}{c}{0.054}    & \multicolumn{1}{c}{0.145}    & \multicolumn{1}{c}{0.047}    & \multicolumn{1}{c}{0.149}    & \multicolumn{1}{c}{0.158}    & \multicolumn{1}{c}{0.036}    &  \multicolumn{1}{c}{0.075}   & \multicolumn{1}{c}{0.099}   \\ 
\hline

\multicolumn{1}{l}{\multirow{2}{*}{\method~(full)}}              
& Best & \multicolumn{1}{c}{\textbf{0.347}}& \multicolumn{1}{c}{\textbf{0.496}}     & \multicolumn{1}{c}{\textbf{0.532}}    & \multicolumn{1}{c}{\textbf{0.332}}    & \multicolumn{1}{c}{\textbf{0.429}}    & \multicolumn{1}{c}{\textbf{0.447}}    & \multicolumn{1}{c}{\textbf{0.504}}    & \multicolumn{1}{c}{\textbf{0.452}}    & \multicolumn{1}{c}{\textbf{0.652}}    & \multicolumn{1}{c}{\textbf{0.531}}    & \multicolumn{1}{c}{\textbf{0.460}}    &  \multicolumn{1}{c}{\textbf{0.652}}   & \multicolumn{1}{c}{\textbf{0.486}}   \\ 
& Last & \multicolumn{1}{c}{\textbf{0.347}}& \multicolumn{1}{c}{\textbf{0.494}}     & \multicolumn{1}{c}{\textbf{0.530}}    & \multicolumn{1}{c}{\textbf{0.329}}    & \multicolumn{1}{c}{\textbf{0.421}}    & \multicolumn{1}{c}{\textbf{0.440}}    & \multicolumn{1}{c}{\textbf{0.502}}    & \multicolumn{1}{c}{\textbf{0.446}}    & \multicolumn{1}{c}{\textbf{0.648}}    & \multicolumn{1}{c}{\textbf{0.531}}    & \multicolumn{1}{c}{\textbf{0.457}}    &  \multicolumn{1}{c}{\textbf{0.650}}   & \multicolumn{1}{c}{\textbf{0.482}}    \\
\hline
\bottomrule
\end{tabular}
}
\caption{The mAP@All retrieval performance comparison for the \method~(full version) and its two variants on OfficeHome dataset. The best performance results among all methods are in bold.}\label{exp_results_ablation}
\end{table*}

\subsection{Implementation Detail}
In \method, a ResNet-50 network pre-trained on ImageNet is utilized to initialize the backbone. Meanwhile, we replace the last FC layer with a 512-D randomly initialized linear layer. The features are $\ell_2$-normalized. To obtain stable performance, four $k$-means are conducted on the obtained features, of which cluster numbers respectively are $\{n_k, 2n_k, 3n_k, 4n_k\}$, where $n_k$ could be empirically set as 50 (for Office31 and ImageCLEF datasets) and 100 (for OfficeHome and Adaptiope datasets). Stochastic Gradient Descent~(SGD) optimizer is adopted to train our \method. For a fair comparison, the hyper-parameters are set as $\eta=0.95$ $n_b=16$, $\lambda=0.01$, $N_e=20$, $\tau=0.01$, and $\alpha=0.003$ for all datasets.
The proposed approach is implemented by PyTorch with two Nvidia GeForce RTX 2080 GPUs. 

\subsection{Experimental Setup}
In the experiments, we evaluate the effectiveness of the proposed approach compared with several state-of-the-art baselines. The compared methods are as follows: 1) \textbf{ImageNet} is a commonly-used baseline that is trained on ImageNet. 2) \textbf{MMD}~\cite{wynne2022mmd} is a kernel-based approach aimed at measuring the distance between two probability distributions in a reproducing kernel Hilbert space.
3) \textbf{InstDis}~\cite{wu2018unsupervised} exploits instance discrimination to achieve unsupervised representation learning. 4) \textbf{SimCLR}~\cite{chen2020simple} learns representations by maximizing agreement between differently augmented views of the same example via a contrastive loss in the latent space. 5) \textbf{CDS}~\cite{kim2021cds} is designed for cross-domain self-supervised pre-training. 6) \textbf{PCS}~\cite{yue2021prototypical} is a cross-domain self-supervised learning method for few-shot unsupervised domain adaptation.
We adopt mean average precision on all retrieved results (mAP@All) as the evaluation metric to measure the performance of the methods. For a fair and comprehensive comparison, we report the best and last mAP@All results among all epochs.

\subsection{Comparison With State-of-the-Art Methods}
We conduct unsupervised cross-domain image retrieval on the four datasets to evaluate the performance of our \method and the compared methods. The experimental results are reported in Tables~\ref{exp_results_office31_adaptiope}, \ref{exp_results_image_clef}, and \ref{exp_results_officehome}. From the experimental results, one could obtain the following observations: 
(1) Our \method outperforms other methods on all datasets, in almost all retrieval cases. The results demonstrate the superiority of the proposed method for unsupervised cross-domain image retrieval. For example, in Table~\ref{exp_results_image_clef} and Table \ref{exp_results_officehome}, our \method respectively surpasses PCS by 1.8\% and 5.9\% in terms of average mAP@All. The reason is that our \method can eliminate the cross-domain gap by encapsulating the discrimination into the domain-invariant embedding space.
(2) Self-supervised representation learning methods are specifically designed for the single-domain task, and cannot achieve satisfactory performance for cross-domain retrieval, \textit{e.g.}, InstDis~\cite{wu2018unsupervised} and SimCLR~\cite{chen2020simple}. The results indicate that the cross-domain gap impedes their performance when applied to multi-domain data.
(3) Compared to self-supervised representation learning methods, CDS~\cite{kim2021cds} and PCS~\cite{yue2021prototypical} achieves better performance. The results indicate that exploiting both in- and cross-domain learning could boost the performance of unsupervised cross-domain image retrieval. 

\subsection{Ablation Study}
In this section, we evaluate the contributions of the proposed components (\textit{i.e.}, $\mathcal{L}_{in}$ and $\mathcal{L}_{cross}$) for unsupervised cross-domain image retrieval.
To this end, we compare our \method with its two variations (\textit{i.e.}, \method with $\mathcal{L}_{in}$ only and \method with $\mathcal{L}_{cross}$ only) on the OfficeHome dataset. The experimental results are shown in Table \ref{exp_results_ablation}. From the table, one could observe that the performance of our \method will degrade significantly in the absence of $\mathcal{L}_{in}$ or $\mathcal{L}_{cross}$, which demonstrates that both the two losses contribute to unsupervised cross-domain image retrieval in our framework. 
Furthermore, the method will fail to work without $L_{in}$, demonstrating that in-domain discrimination excavation is crucial for cross-domain retrieval.

\subsection{Effect of Coefficient $\lambda$}
\begin{figure}[t]
     \centering
     \begin{subfigure}[b]{0.23\textwidth}
         \centering
         \includegraphics[width=\textwidth]{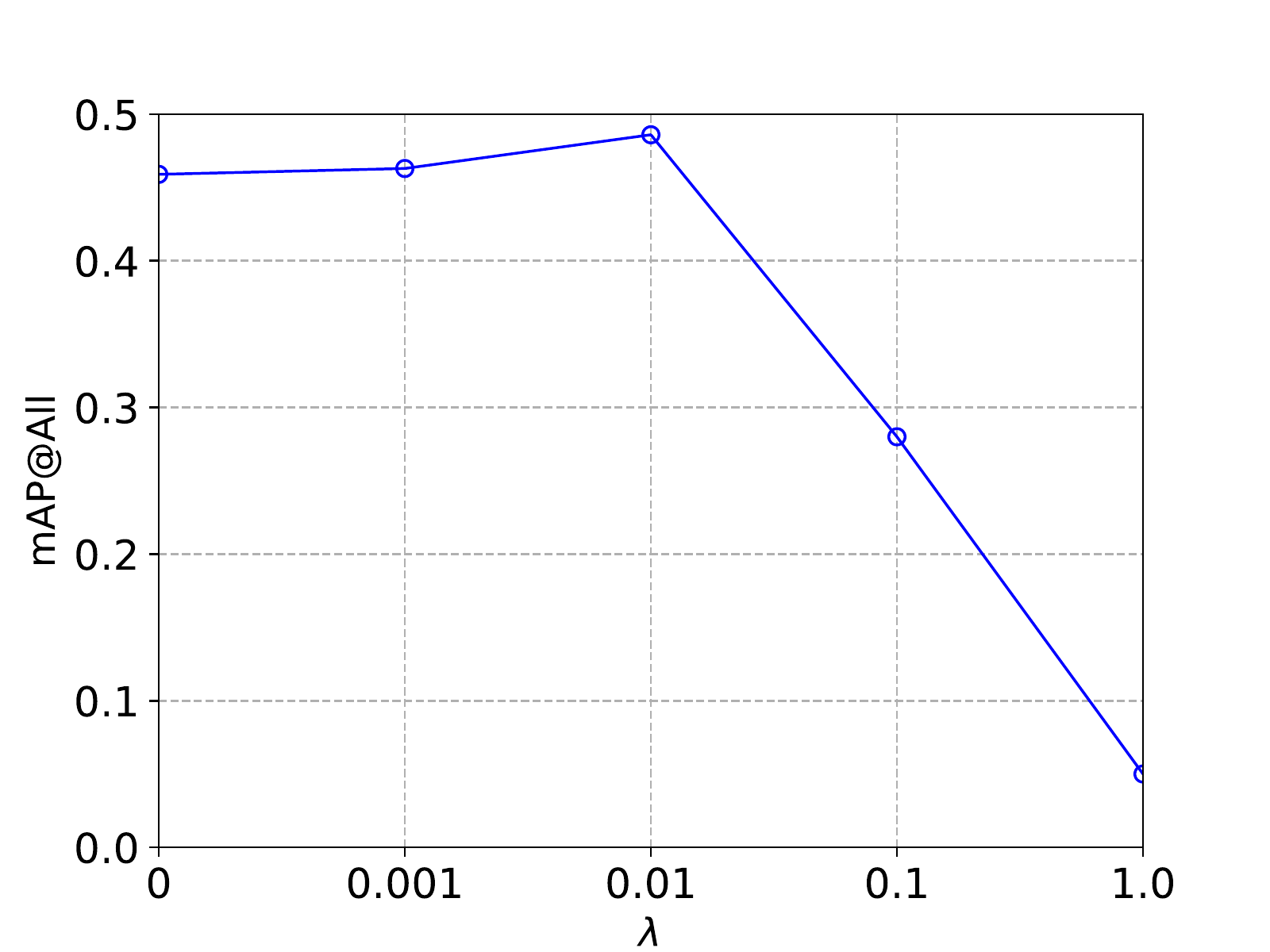}
         \caption{OfficeHome}
     \end{subfigure}
     \hfill
     \begin{subfigure}[b]{0.23\textwidth}
         \centering
         \includegraphics[width=\textwidth]{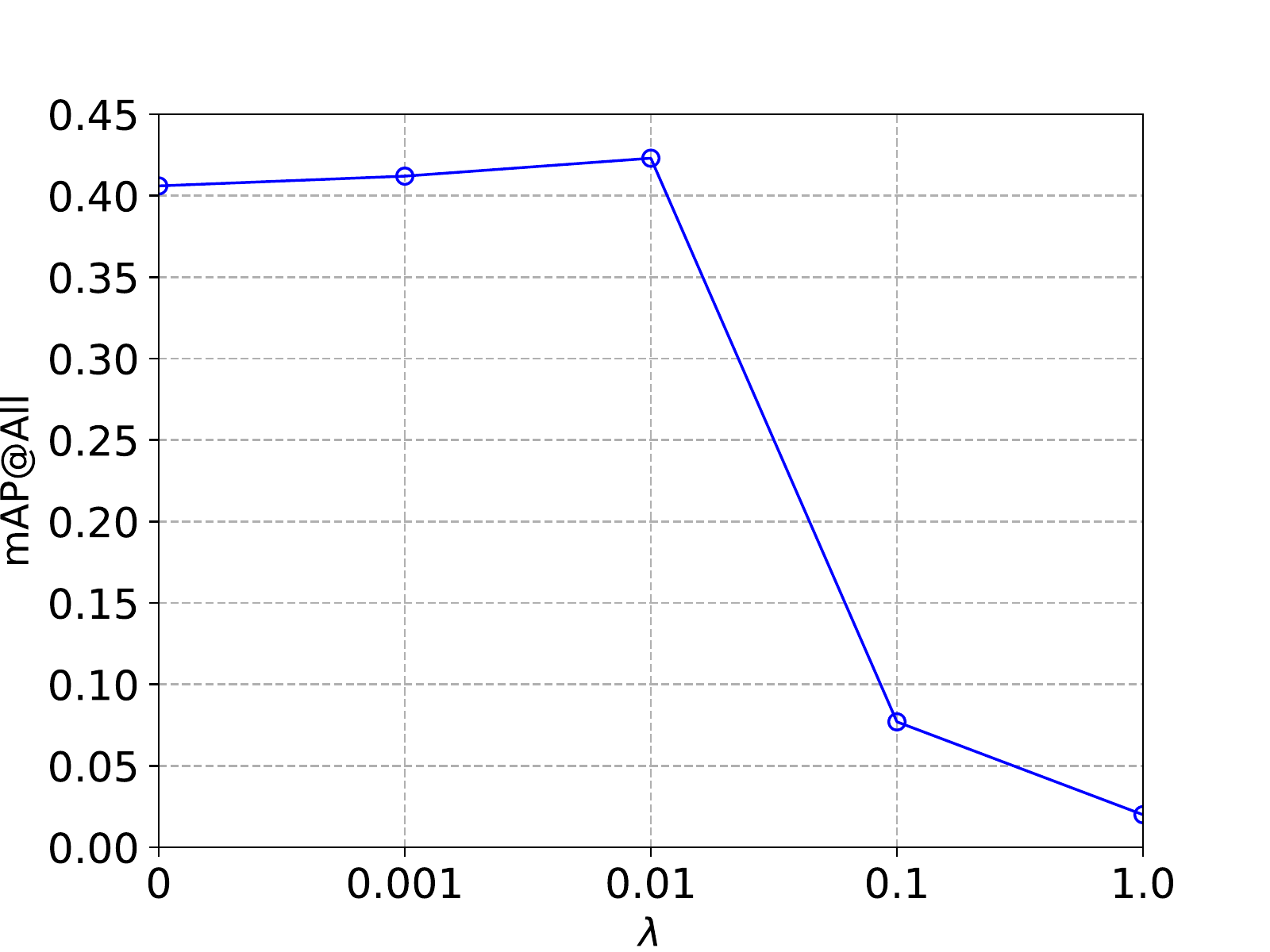}
         \caption{Adaptiope}
     \end{subfigure}
    \caption{Unsupervised cross-domain image retrieval performance of \method in terms of mAP@All scores versus different values of $\lambda$ on OfficeHome and Adaptiope datasets.}
    \label{fig:coefficient_curve}
\end{figure}

To investigate the impact of the coefficient $\lambda$ in Eq.~(\ref{loss}), we conduct parameter analysis experiments on OfficeHome and Adaptiope as shown in Figure \ref{fig:coefficient_curve}. The figure plots mAP@all scores \textit{w.r.t.} different values of $\lambda$. In the figure, one could find that setting $\lambda=0.01$ achieves the best performance on both OfficeHome and Adaptiope. Based on the observation, we set $\lambda=0.01$ in all experiments.

\begin{figure}[!h]
     \centering
     \begin{subfigure}[b]{0.21\textwidth}
         \centering
         \includegraphics[width=\textwidth]{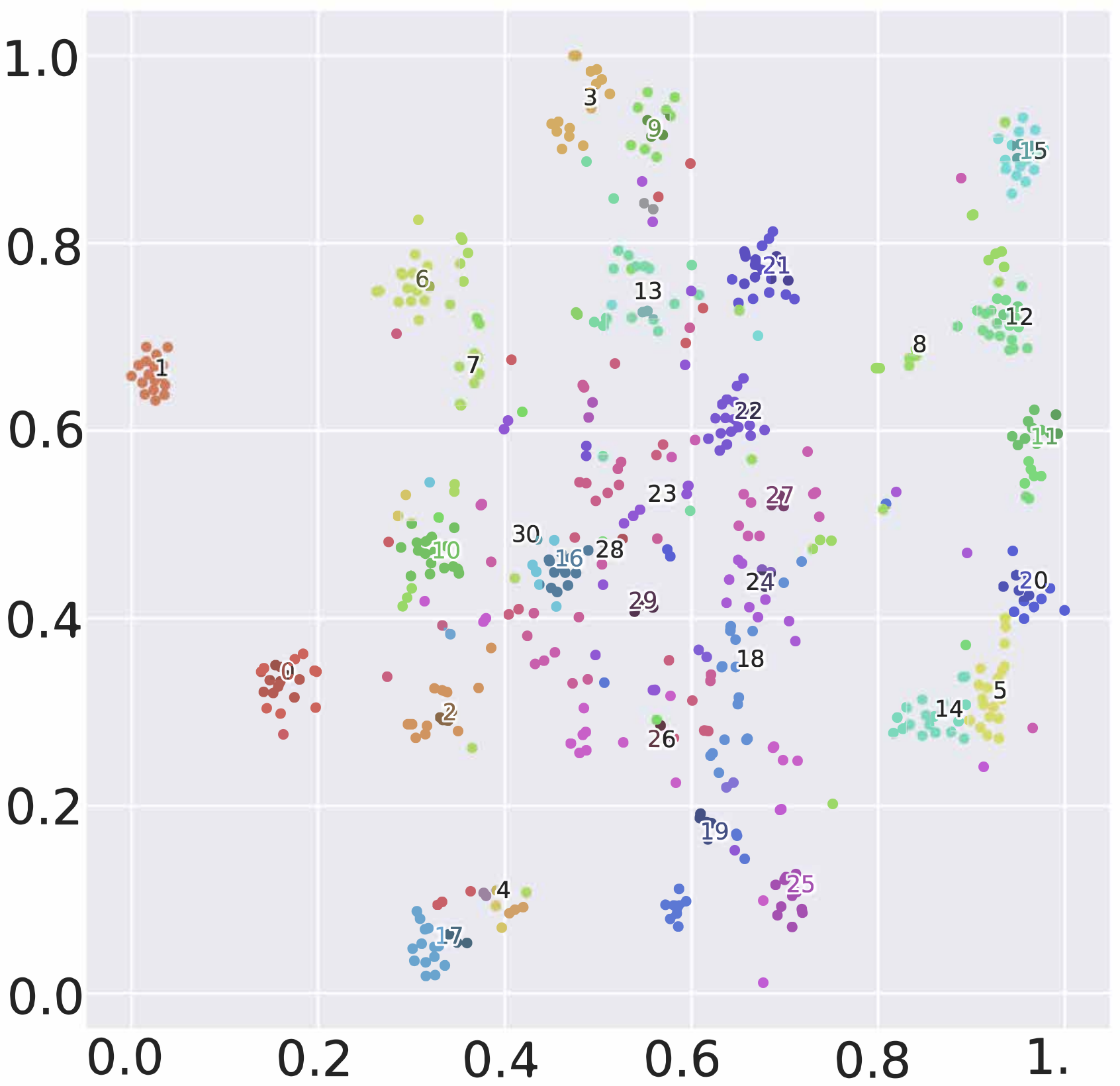}
         \caption{Amazon~(ImageNet)}
     \end{subfigure}
     \hfill
     \begin{subfigure}[b]{0.21\textwidth}
         \centering
         \includegraphics[width=\textwidth]{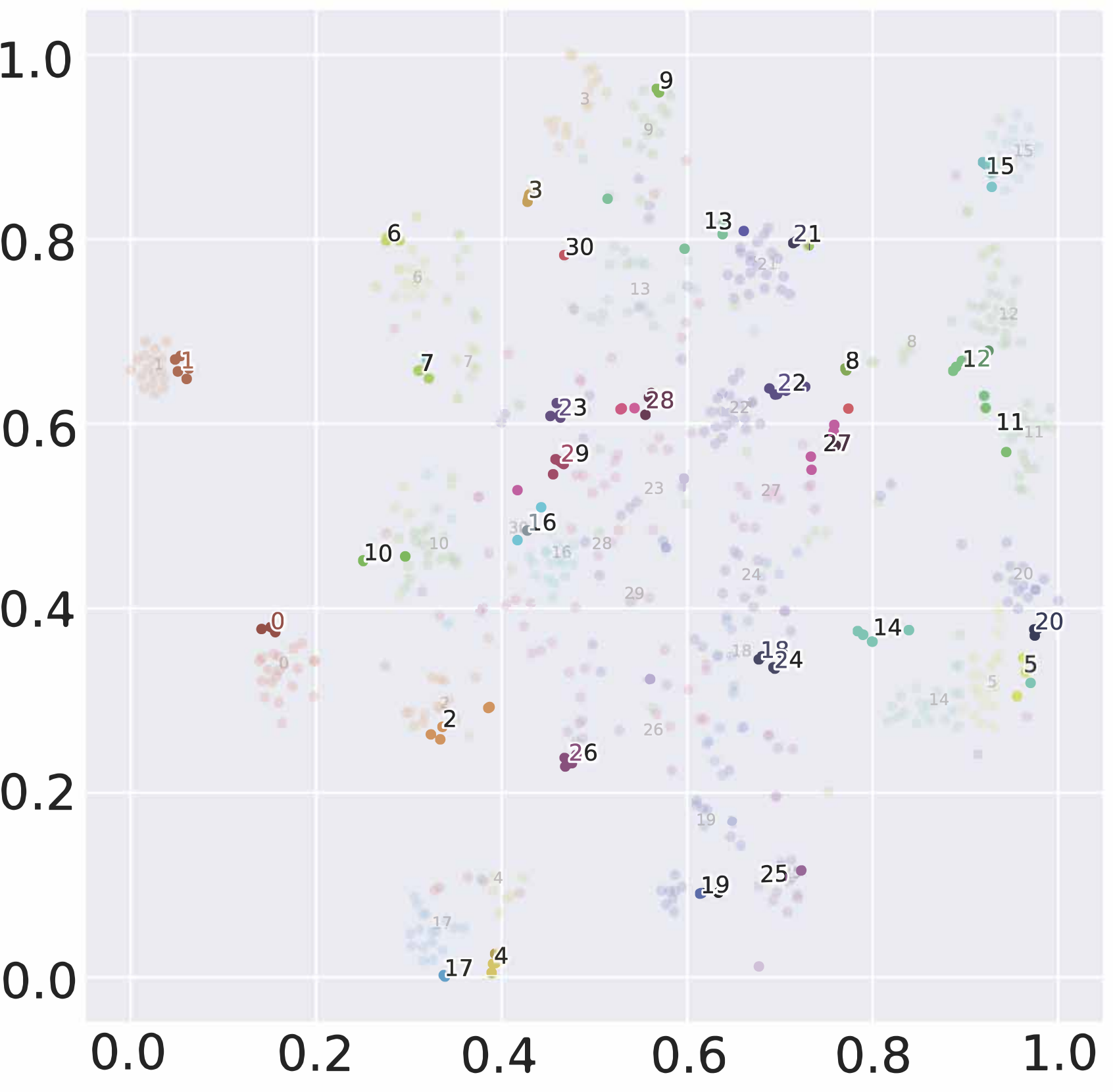}
         \caption{DSLR~(ImageNet)}
     \end{subfigure}\\
     \begin{subfigure}[b]{0.21\textwidth}
         \centering
         \includegraphics[width=\textwidth]{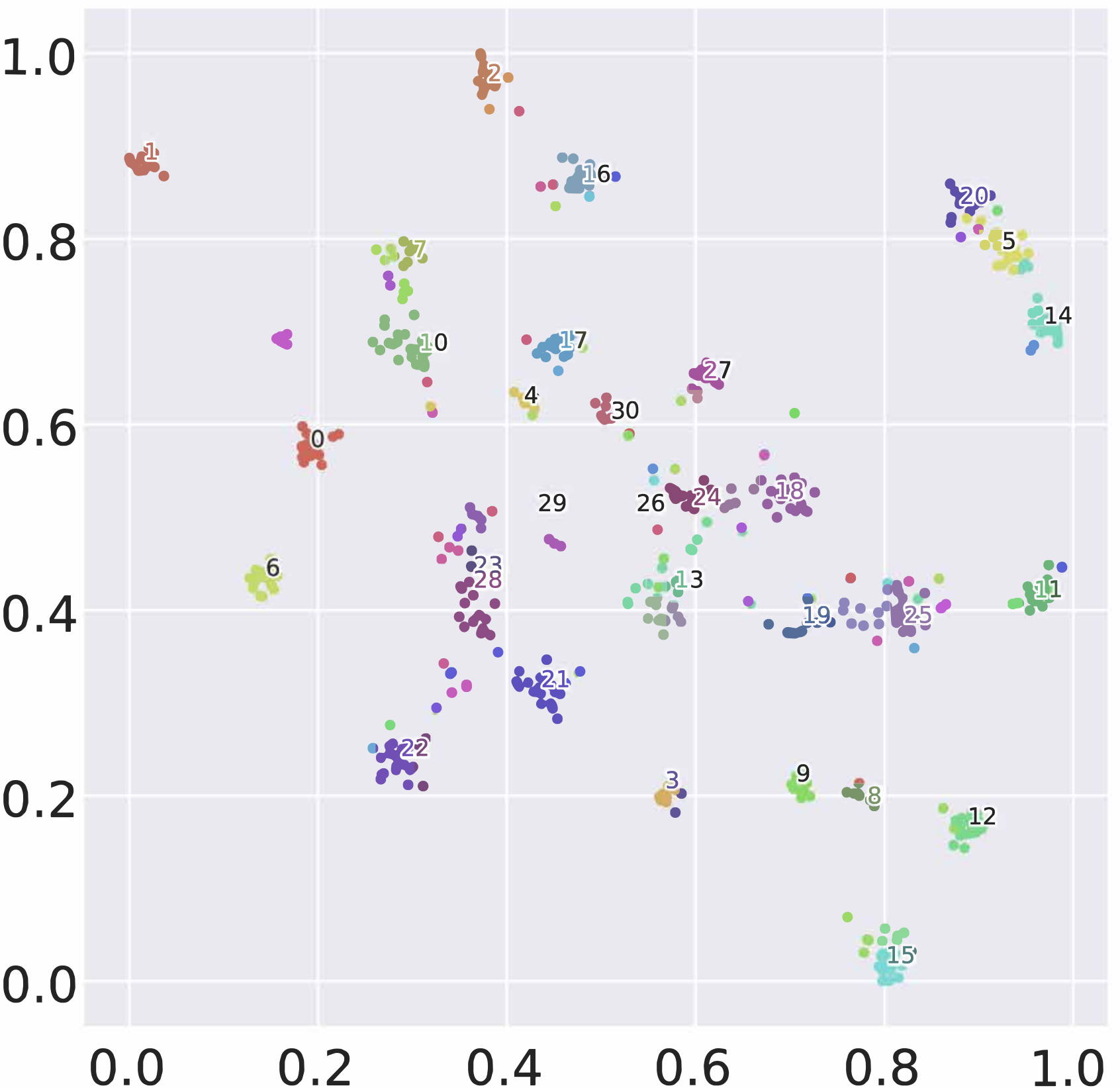}
         \caption{Amazon~(CDS)}
     \end{subfigure}
     \hfill
     \begin{subfigure}[b]{0.21\textwidth}
         \centering
         \includegraphics[width=\textwidth]{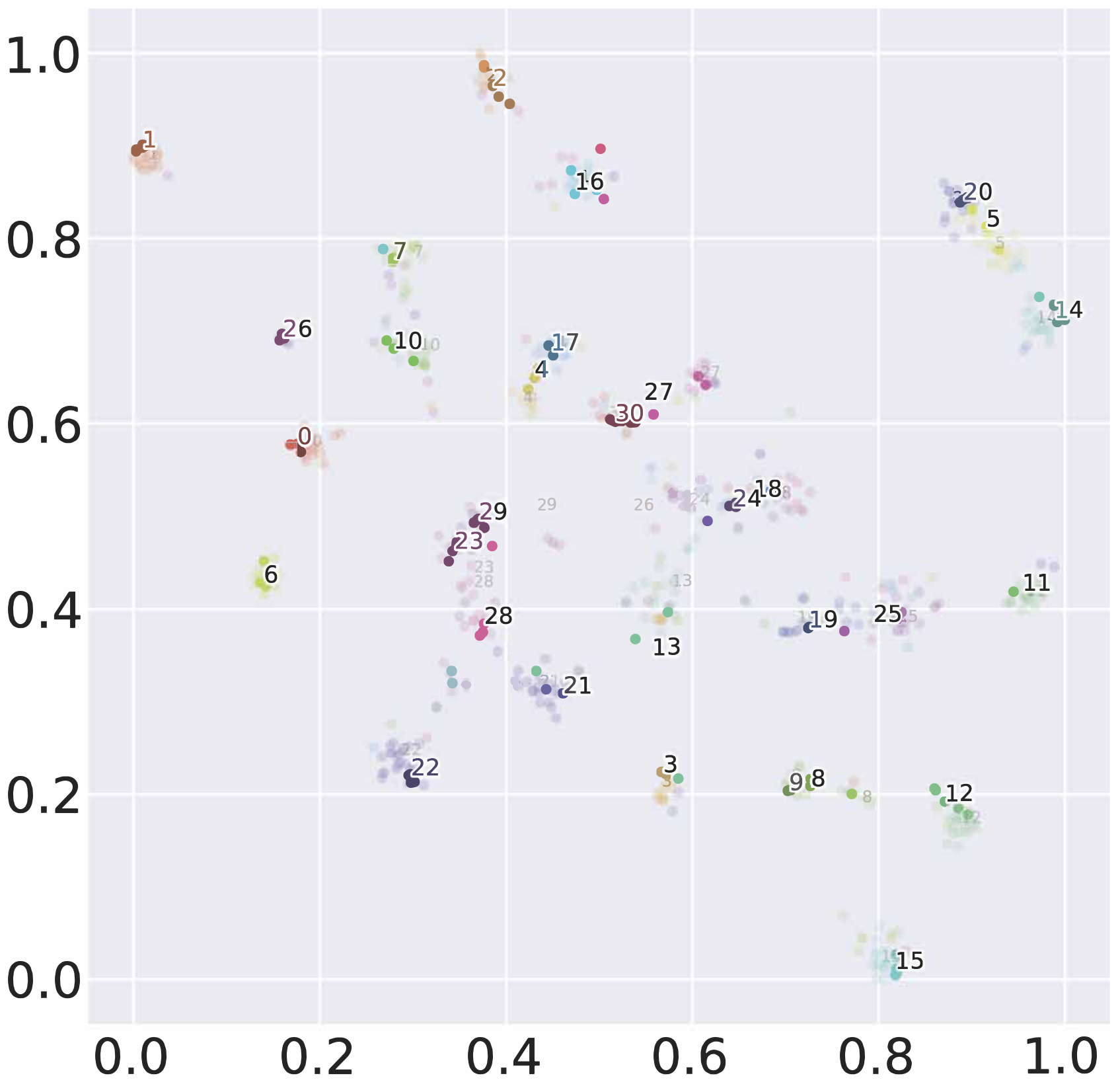}
         \caption{DSLR~(CDS)}
     \end{subfigure}\\
      \begin{subfigure}[b]{0.21\textwidth}
         \centering
         \includegraphics[width=\textwidth]{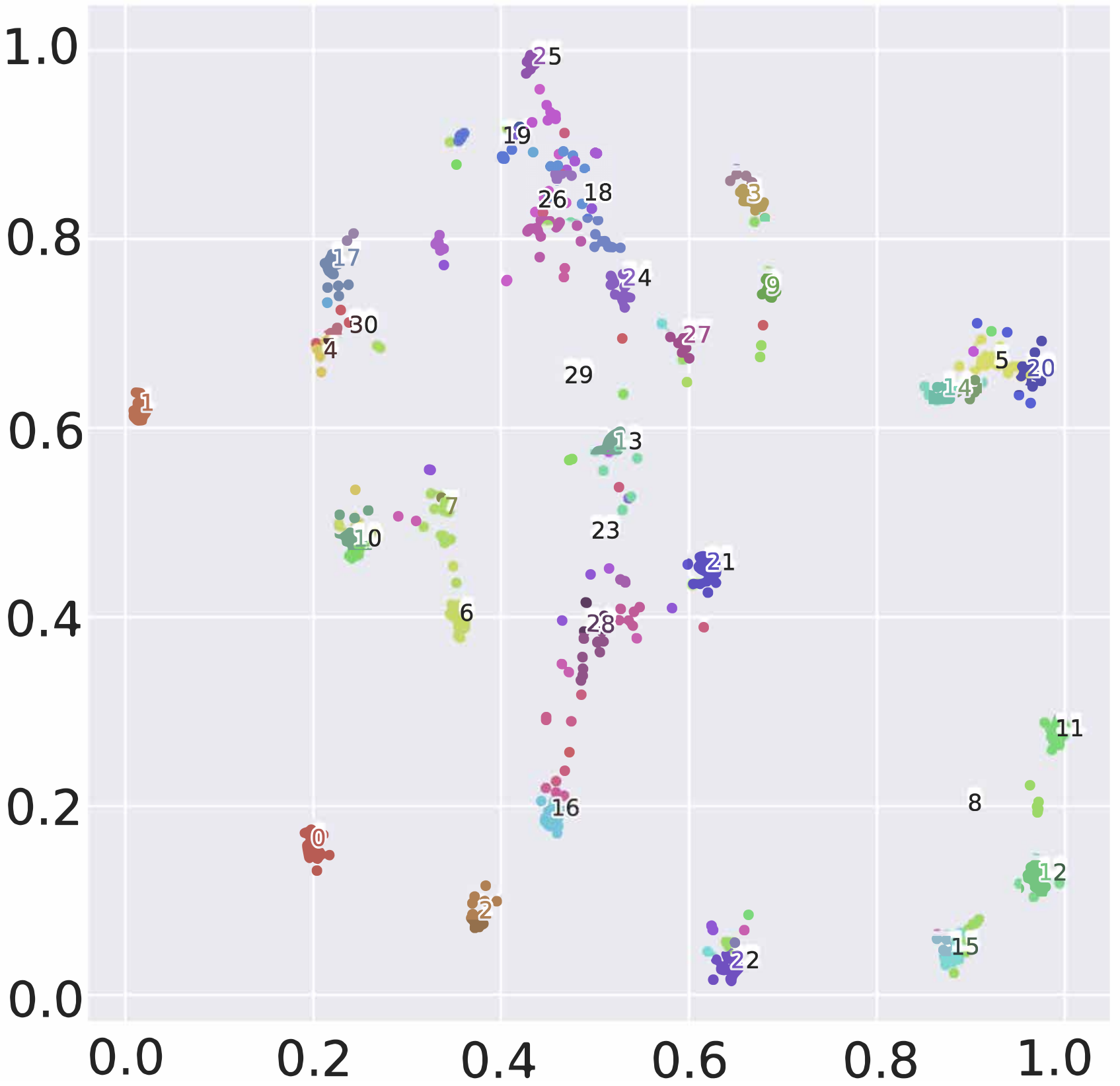}
         \caption{Amazon~(\method)}
     \end{subfigure}
     \hfill
     \begin{subfigure}[b]{0.21\textwidth}
         \centering
         \includegraphics[width=\textwidth]{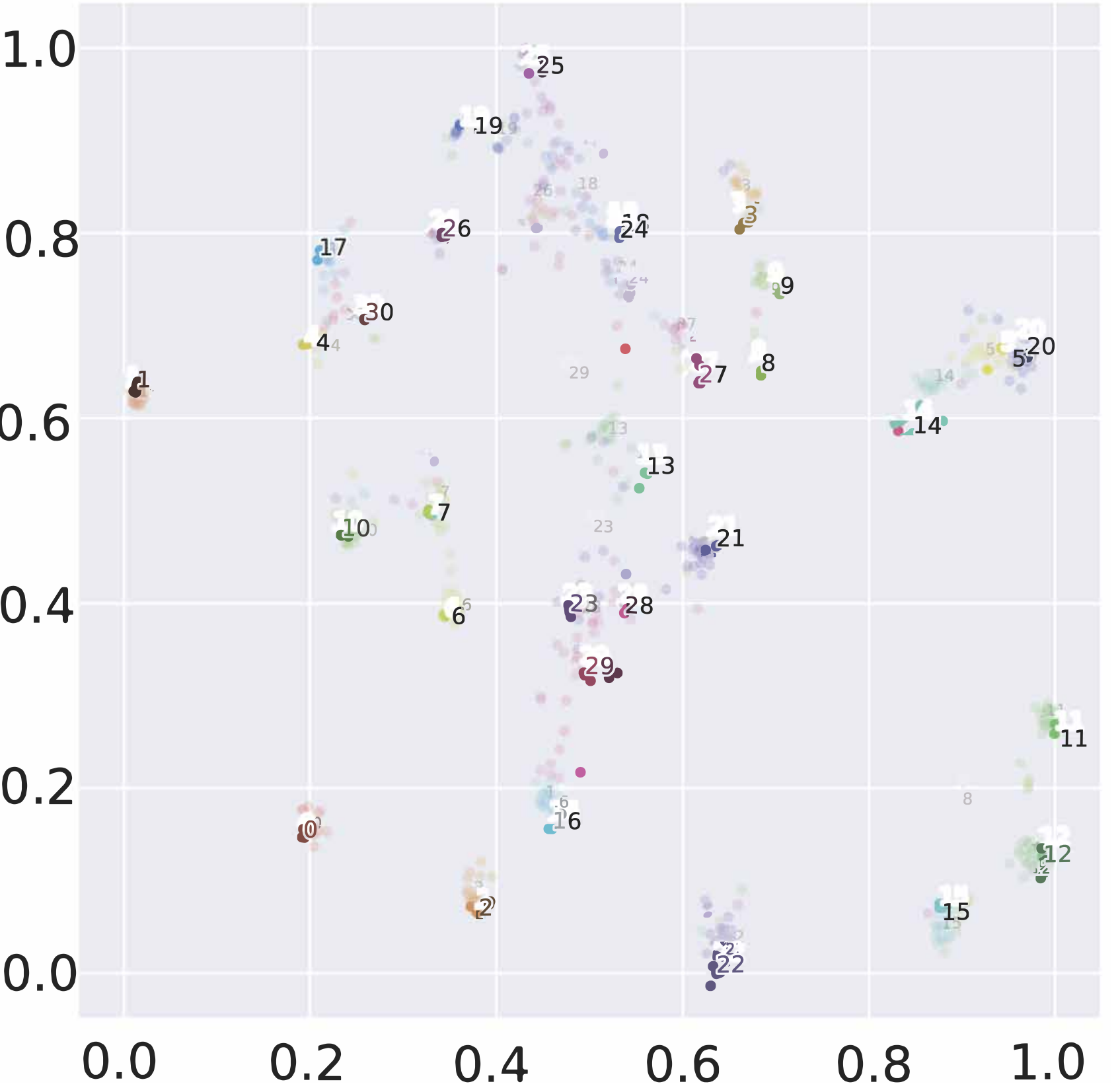}
         \caption{DSLR~(\method)}
     \end{subfigure}
    \caption{2-d t-SNE visualizations using 512-d feature representations learned by ImageNet pre-pretraining, CDS, and the proposed \method on the testing set of Office31. Each sample is represented by a marker and colored by its corresponding label. \textit{Best viewed in color.}}
    \label{fig:tsne}
\end{figure}
\subsection{Visualization of the Learned Embeddings}
To demonstrate the discriminative information of the learned embeddings, we plot the learned embeddings of ImageNet-pretraining, CDS, and \method with t-SNE~\cite{van2008tsne} on the Amazon-to-DSLR setting in Office31. The illustrations are shown in Figure~\ref{fig:tsne}. From the figure, one could observe that: (1) Compared with ImageNet-pretraining and CDS, the proposed approach well clusters the features with the same class from both domains, demonstrating that our \method favors more discriminative features. (2) The features of \method from the two domains are well aggregated, which demonstrates that \method learns better domain-alignment for different domains.

\subsection{Example of Retrievals} 
\begin{figure}[t]
	\centering
\includegraphics[width=0.45\textwidth]{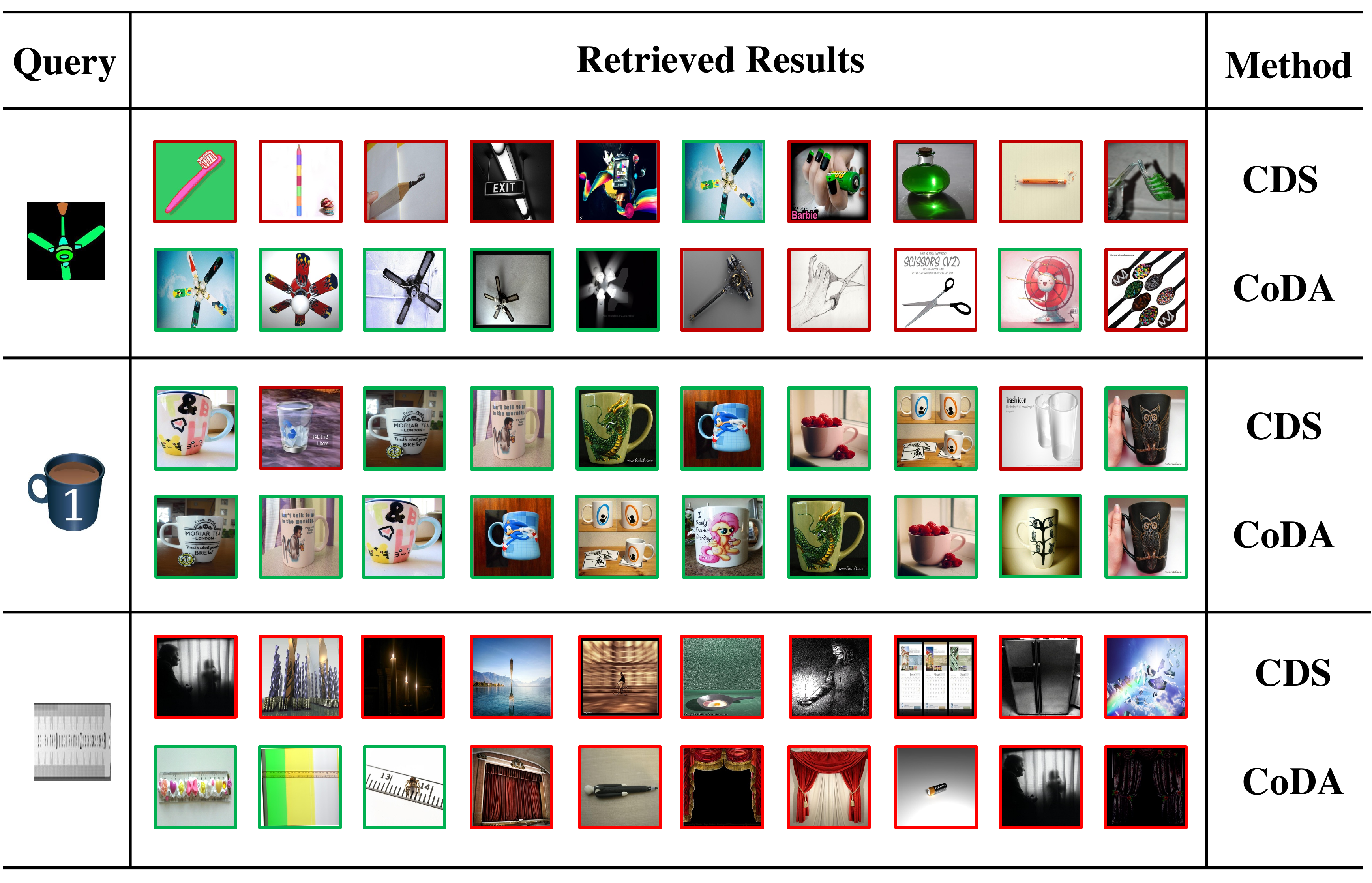}
	\captionsetup{font={small}}
	\caption{Top-10 UCIR results obtained by CDS and \method on OfficeHome dataset. Retrieval is performed by nearest neighbors search using cosine distance on 512-d real value feature vectors. Green borders denote correctly retrieved results, and the red borders demonstrate incorrect retrieved candidates.}\label{fig:officehome_qualitative_results}
\end{figure}

In Figure~\ref{fig:officehome_qualitative_results}, we visually show the top 10 unsupervised cross-domain image retrieval results using CDS and \method approaches. The retrievals are conducted on the OfficeHome with 512-D real value features for task C-A~(Clipart-Art). The red borders indicate wrongly retrieved results while the green borders denote correctly retrieved results. From these examples, we can observe that the proposed method obtains promising results in most cases compared to CDS. For example, in the third query, given the ``ruler'' query, CDS fails to retrieve correct images, while our \method could find the correct ones in top-3 retrievals. 
In the wrongly retrieved results, \method fails to search for the correct images since the queries and the retrieved images are visually similar. For instance, given the ``fan'' image (in the first query), a ``scissor'' image is wrongly retrieved, which shares some visual similarities with the ``fan'' image.

\section{Conclusion}
This paper proposes a novel method, called \method for unsupervised cross-domain image retrieval, to project different domains into a common space. The proposed method is designed to bridge the domain gap through In-domain Self-matching Supervision (ISS) and Cross-domain Classifier Alignment (CCA). Specifically, ISS encapsulates discriminative information into common representations by self-matching. Meanwhile, CCA aligns the features from different domains by minimizing the discrepancy among different domain-specific classifiers. Thanks to ISS and CCA, \method enjoys more discriminability in the common space and embraces the domain invariance in unsupervised cross-domain image retrieval. Comprehensive experimental results on four benchmarks have verified the effectiveness of \method.

\section{Acknowledgments}
This work is supported by the National Natural Science Foundation of China (Grant Nos. U19A2078, 61971296, 62102274), China Postdoctoral Science Foundation (Nos. 2021TQ0223, 2022M712236, 2021M692270), Sichuan Science and Technology Planning Project (Grant Nos. 2023ZHCG0016, 2023YFG0033, 2022YFQ0014, 2022YFH0021, 2021YFS0389, 2021YFS0390), and Fundamental Research Funds for the Central Universities (No. 2022SCU12081).

\bibliography{coda}




\end{document}